\documentclass[runningheads]{llncs}
\usepackage[a4paper,margin=25mm,headheight=12pt,headsep=16pt]{geometry}
\usepackage[utf8]{inputenc}
\usepackage[T1]{fontenc}
\usepackage{newtxtext}
\usepackage{amsmath}
\usepackage[varvw]{newtxmath}
\usepackage{graphicx}
\usepackage{hyperref}
\usepackage{cleveref}
\hypersetup{hidelinks}
\newcommand{\methodname}{EchoChat}

\newcommand{\bn}[1]{#1}
\crefname{figure}{Fig.}{Figs.}
\Crefname{figure}{Fig.}{Figs.}
\begin{document}
\title{Multimodal Generative AI Assists Comprehensive Echocardiographic Interpretation}
\titlerunning{Multimodal AI for Echocardiographic Interpretation}
\author{Yi Qin\inst{1,\dagger} \and
Marawan Elbatel\inst{1,\dagger} \and
Shanshan Song\inst{1,\dagger} \and
Taoran Huang\inst{2,\dagger} \and
Jiarong Guo\inst{1} \and
Ningshu Li\inst{1} \and
Lehan Wang\inst{1} \and
Chenxu Zhao\inst{3} \and
Jiexin He\inst{3} \and
Machteld J. Boonstra\inst{4} \and
Stephan van der Zwaard\inst{4} \and
Marielle C. van de Veerdonk\inst{4} \and
Pere Lopez-Gutierrez\inst{5,6} \and
Lydia Dux-Santoy\inst{5} \and
Andrea Guala\inst{5,7} \and
Enrique Almar-Munoz\inst{8} \and
Sebastian Johannes Reinstadler\inst{9} \and
Radek Debiec\inst{10,11,12,13} \and
Yi-Hang Zhu\inst{14} \and
Huiyu Zhou\inst{14} \and
Adrian P Cheong\inst{15} \and
Yike Guo\inst{1,16} \and
Agnes Mayr\inst{8} \and
Karim Lekadir\inst{17} \and
Yong Jiang\inst{21} \and
Nilesh J. Samani\inst{10,11,12,13} \and
Dacheng Tao\inst{18} \and
Folkert W. Asselbergs\inst{4,19,20} \and
Hongwen Fei\inst{2} \and
Alex PW Lee\inst{3} \and
Xiaomeng Li\inst{1,16,22}}
\authorrunning{Y. Qin et al.}
\institute{Department of Electronic and Computer Engineering, The Hong Kong University of Science and Technology, Hong Kong SAR, China \and
Guangdong Cardiovascular Institute, Guangdong Provincial People’s Hospital (Guangdong Academy of Medical Sciences), Southern Medical University, Guangzhou, Guangdong Province, China \and
Li Ka Shing Institute of Health Sciences, Faculty of Medicine, The Chinese University of Hong Kong, Hong Kong SAR, China \and
Department of Cardiology, Amsterdam Cardiovascular Sciences, Amsterdam University Medical Center, University of Amsterdam, Amsterdam, the Netherlands \and
Vall d’Hebron Institut de Recerca, Barcelona, Spain \and
Universitat Autònoma de Barcelona, Bellaterra, Spain \and
CIBER-CV, Instituto de Salud Carlos III, Madrid, Spain \and
Department of Radiology, Medical University of Innsbruck, Innsbruck, Austria. \and
Department of Internal Medicine III, Medical University of Innsbruck, Innsbruck, Austria \and
Division of Cardiovascular Sciences, University of Leicester, Leicester, United Kingdom \and
British Heart Foundation Centre of Research Excellence, University of Leicester, Leicester, United Kingdom \and
NIHR Leicester Biomedical Research Centre, University of Leicester, Leicester, United Kingdom \and
Centre for Digital Health and Precision Medicine, University of Leicester, Leicester, United Kingdom \and
School of Computing and Mathematical Sciences, University of Leicester, Leicester, United Kingdom \and
The Heart Clinic (Central), Virtus Medical Group, Hong Kong SAR, China \and
Department of Computer Science and Engineering, The Hong Kong University of Science and Technology, Hong Kong SAR, China \and
Institució Catalana de Recerca i Estudis Avançats (ICREA), Artificial Intelligence in Medicine Lab (BCN-AIM), Universitat de Barcelona, Spain \and
College of Computing and Data Science, Nanyang Technological University, Singapore \and
The National Institute for Health Research University College London Hospitals Biomedical Research Centre, University College London, London, United Kingdom \and
Institute of Health Informatics, University College London, London, United Kingdom \and
Fuwai Shenzhen Hospital, Shenzhen, China \and
Shenzhen Loop Area Institute, Shenzhen, China}
\maketitle
\begin{center}
\footnotesize $\dagger$ These authors contributed equally to this work.
\end{center}
\begin{abstract}
\noindent \bn{Echocardiography is the cornerstone of cardiovascular diagnostics, yet its global utility is constrained by a critical shortage of experienced interpreters.} Current artificial intelligence (AI) solutions remain limited, typically operating as opaque ``black boxes'' or focusing on narrow tasks that fail to capture the holistic, guideline-driven reasoning of a clinician. Here we introduce EchoChat, a generative multimodal vision-language model that performs comprehensive echocardiographic analysis to support clinician interpretation. By emulating the cognitive workflow of a cardiologist, EchoChat integrates vast medical knowledge with a chain-of-thought reasoning mechanism to explicitly ground its conclusions in visual evidence. \bn{Trained on echocardiography-instruction pairs from 46,358 examinations, our model establishes a new state-of-the-art across diverse cohorts from six international medical centers. It supports the broadest diagnostic scope of 33 cardiac conditions and findings, including 15 newly added conditions for which it achieves an average F1 of 73.7\% across available external evaluations. It surpasses existing best-performing foundation models in cardiac condition recognition (improving average recall by 13.54\% on shared tasks) and quantification (average $R^2$ score increase of 0.275 on shared tasks). Furthermore, in a source-blinded echocardiographer reader study, echocardiography reports co-generated with EchoChat were rated higher in diagnostic accuracy and completeness by two senior echocardiographers, and improved diagnostic performance from 72.78\% to 82.75\% in terms of F1 score over routine-reader only reports.} These results demonstrate that integrating clinical reasoning into multimodal AI can democratize specialist expertise and transform the standard of cardiovascular care.
\end{abstract}

\section{Introduction}

Cardiovascular diseases remain the leading cause of mortality worldwide, necessitating scalable solutions for accurate and reliable diagnostics~\cite{global2025global,martin20252025,firima2024implementing}. Echocardiography is central to this clinical challenge, providing a non-invasive, real-time window into cardiac structure and function. However, the life-saving potential of this technology is severely constrained by a global shortage of experienced interpreters~\cite{american2023workforce,patel2025cardiology}. Achieving proficiency in echocardiography requires years of intensive training, creating a bottleneck that restricts access to timely diagnosis, particularly in resource-limited settings~\cite{muriuki2025cardiovascular,popescu2009european,wiegers20192019}. While Artificial Intelligence (AI) has emerged as a promising solution, current approaches remain fragmented. Existing tools are either task-specific models limited to predicting individual diseases~\cite{marangou2019echocardiography,huson2019cardiac,justin2024routine,ouyang2020video,sahashi2025artificial}, or general-purpose foundation models that perform isolated measurements and diagnosis without clinical context~\cite{holste2025complete,vukadinovic2026comprehensive}. Crucially, these systems operate as ``black boxes,'' failing to replicate the holistic, guideline-driven reasoning process of a cardiologist. Consequently, a critical gap remains: the absence of an AI system capable of comprehensive, trustworthy, and explainable echocardiographic analysis assistive to the echocardiographer.

\begin{figure}[!htbp]

    \centering
    \includegraphics[width=\textwidth,height=0.70\textheight,keepaspectratio]{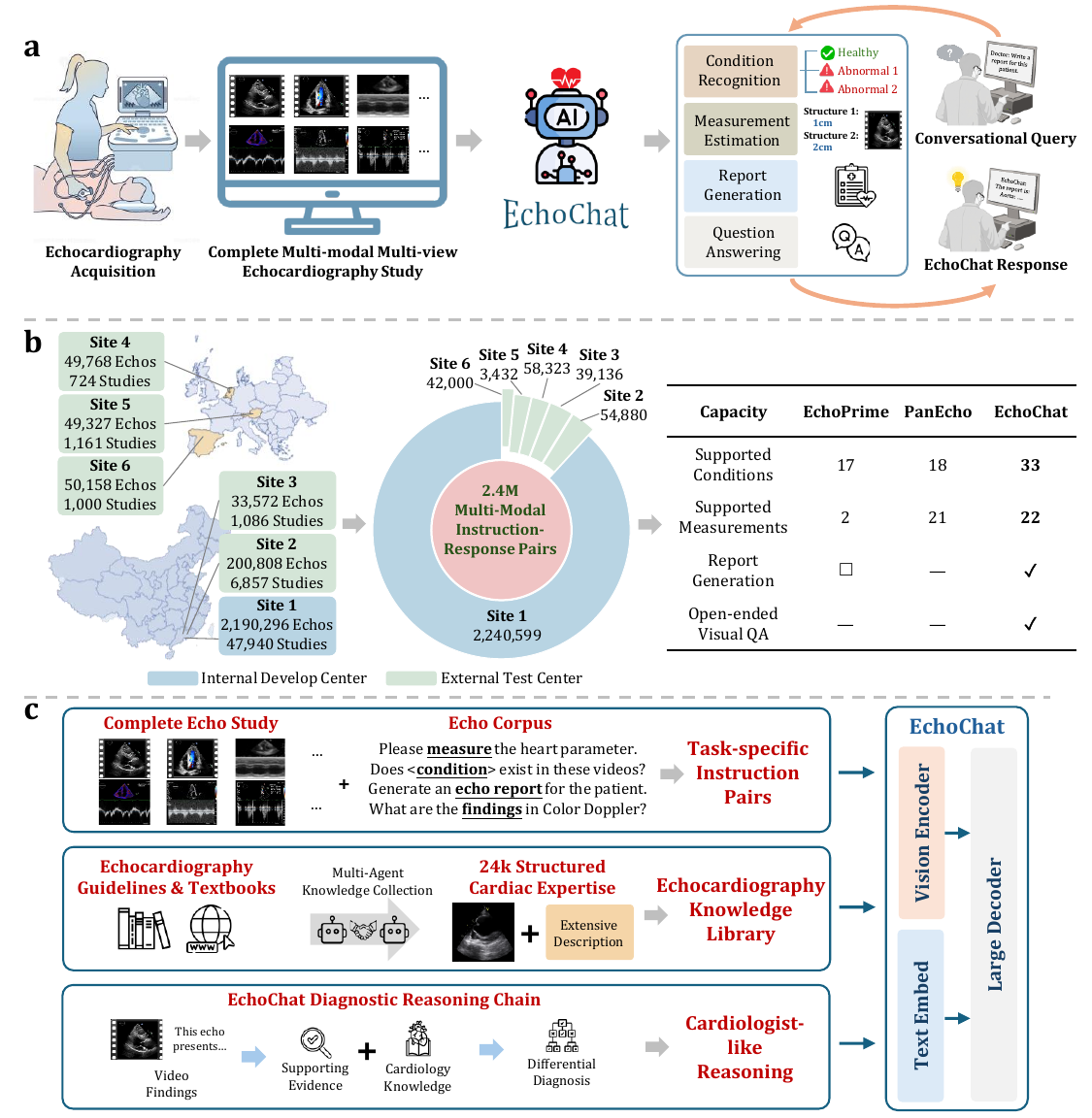}
    \caption{\textbf{The overall workflow of \methodname{}.}
    \textbf{a,} \methodname{} streamlines the clinical workflow by functioning as an interactive assistant that enables conversational queries. The system processes these queries to execute comprehensive echocardiographic tasks, thereby supporting the echocardiographer's decision-making.
    \textbf{b,} \methodname{} is a multi-modal large language model trained and tested on over 2 million echocardiographic videos and images from six international hospitals. By integrating imaging and textual data, it performs recognition on 33 types of cardiac conditions and findings, 22 types of measurement estimation, report generation, and visual question answering, outperforming existing echocardiographic foundation models with new capabilities. $\Box$ refers to the report retrieval capacity of EchoPrime, which matches existing reports to echocardiographic content based on similarity rather than generating a new report from scratch.
    \textbf{c,} By employing echocardiography-centered, evidence-based reasoning alongside a vast knowledge library in the training corpus, \methodname{} enhances disease recognition accuracy and provides interpretable supporting evidence.
    }
     \label{fig:intro}

\end{figure}

Here we introduce EchoChat, a generative multimodal large vision-language model that bridges this gap, \bn{functioning as an AI assistant designed to help echocardiographers interpret echocardiography, estimate cardiac measurements and draft reports}. EchoChat represents a paradigm shift from narrow, task-specific algorithms to a system capable of executing complete clinical diagnostic analysis. \bn{It performs a comprehensive analysis covering an unprecedented range of 33 cardiac conditions and 22 cardiac measurements. Unlike previous iterations of medical AI, EchoChat generates complete, narrative diagnostic reports from raw echocardiography data and engages in evidence-based question-answering regarding its findings (\Cref{fig:intro}a).} We achieved this by developing a framework that emulates human expert learning and reasoning, integrating two core innovations as illustrated in~\Cref{fig:intro}c: an Echocardiography Knowledge Library (EKL), which grounds the model in a vast database of established cardiological science; and a Cardiologist-Like Reasoning (CLR) module, a bespoke Chain-of-Thought mechanism that enforces adherence to clinical guidelines and explicitly links conclusions to visual evidence.

Trained on echocardiography-instruction pairs derived from \bn{46,358 echocardiographic examinations} (\Cref{fig:intro}b), EchoChat establishes a new state-of-the-art.
In evaluations involving over 5,000 echocardiography studies across five external hospitals in the Netherlands, Austria, Spain, and China, EchoChat demonstrates enhanced diagnostic scope and performance, as well as generalization across diverse populations and acquisition protocols. Beyond the coverage of existing foundation models (18 conditions at most~\cite{holste2025complete}), \bn{EchoChat extends diagnostic scope to a broad set of 33 conditions with 15 newly available challenging conditions achieving an average accuracy of 83.0\% across four external test sets. Furthermore, EchoChat surpasses leading foundation models~\cite{holste2025complete,vukadinovic2026comprehensive} in both cardiac condition recognition (average recall improvement of 28.00\% and 13.54\% over EchoPrime~\cite{vukadinovic2026comprehensive} and PanEcho~\cite{holste2025complete}, respectively) and quantification (average $R^2$ score increase of 0.294 and 0.275 over EchoPrime~\cite{vukadinovic2026comprehensive} and PanEcho~\cite{holste2025complete}, respectively). Notably, EchoChat achieved a mean absolute error in left ventricular ejection fraction estimation that was 1.79\% lower than that of PanEcho~\cite{holste2025complete}, averaged equally across the five external clinical cohorts.
In the EchoChat-routine echocardiographer collaboration study, EchoChat-routine echocardiographer collaborative reports improve average primary-diagnosis F1 scores from 72.78\% to 82.75\% compared with the unassisted echocardiographer group.
Two senior echocardiographers blinded to report source also rated collaborative reports higher in diagnostic accuracy and completeness, with mean scores 15.84\% and 20.74\% higher than those of unassisted routine reports, respectively.
}
By democratizing access to high-level cardiac expertise and augmenting clinician performance, EchoChat offers a transformative path toward a new standard of global cardiovascular care.

\section{Results}

\methodname{} is developed and validated on large-scale patient cohorts consisting of \bn{2,573,929 echocardiographic videos and images} within \bn{58,768 echocardiography studies} (echocardiography reports and measurements) from six international centers (\Cref{fig:intro}b). The development data consists of multiple cardiac views and echocardiographic modalities, covering the majority of data types in a routine echocardiography study. Of these studies, \bn{46,358 echocardiographic studies} from Hong Kong SAR China formed the training set. \bn{Moreover, 5,095 echocardiography studies from five other hospitals from Mainland A China (1,124 studies), Mainland B China (1,086 studies), Spain (1,000 studies), Austria (1,161 studies), and the Netherlands (724 studies) formed five external test sets}. Mainland A contributed 6,857 examinations in total, including its 1,124 external-test examinations. Of the 6,857, 100 were held out for the reader study and the rest of 6,757 were used for Chinese report-adapter training; all external evaluations used the model without that adapter.

\methodname{} is an echocardiography multi-modal large language model (MLLM). It takes a complete echocardiography study, including recorded videos and key frames, and the echocardiography-related instructions as the input of the model, and outputs the corresponding textual and numerical responses. In a supervised approach, \methodname{} is trained on multi-modality instruction tuning dataset constructed based on the curated echocardiography studies using cross-entropy loss. \methodname{} utilized Qwen 2.5-VL MLLM's vision encoder and textual decoder as the base model for development, leveraging the model's strong performance in multimodal understanding and natural language generation.

\subsection{\methodname{} integrates condition recognition and quantitative measurement across major domains of echocardiography}

We first evaluated \methodname{}'s performance on cardiac condition recognition and heart measurement estimation tasks in our internal held-out test set. EchoChat was evaluated on 33 cardiac conditions and findings across six clinical categories: valvular abnormalities; ventricular abnormalities and cardiomyopathies; atrial abnormalities; vascular and pulmonary circulatory abnormalities; device-related and post-intervention states; and pericardial abnormalities.

\begin{figure}[!htbp]

    \centering
    \includegraphics[width=\textwidth,height=0.65\textheight,keepaspectratio]{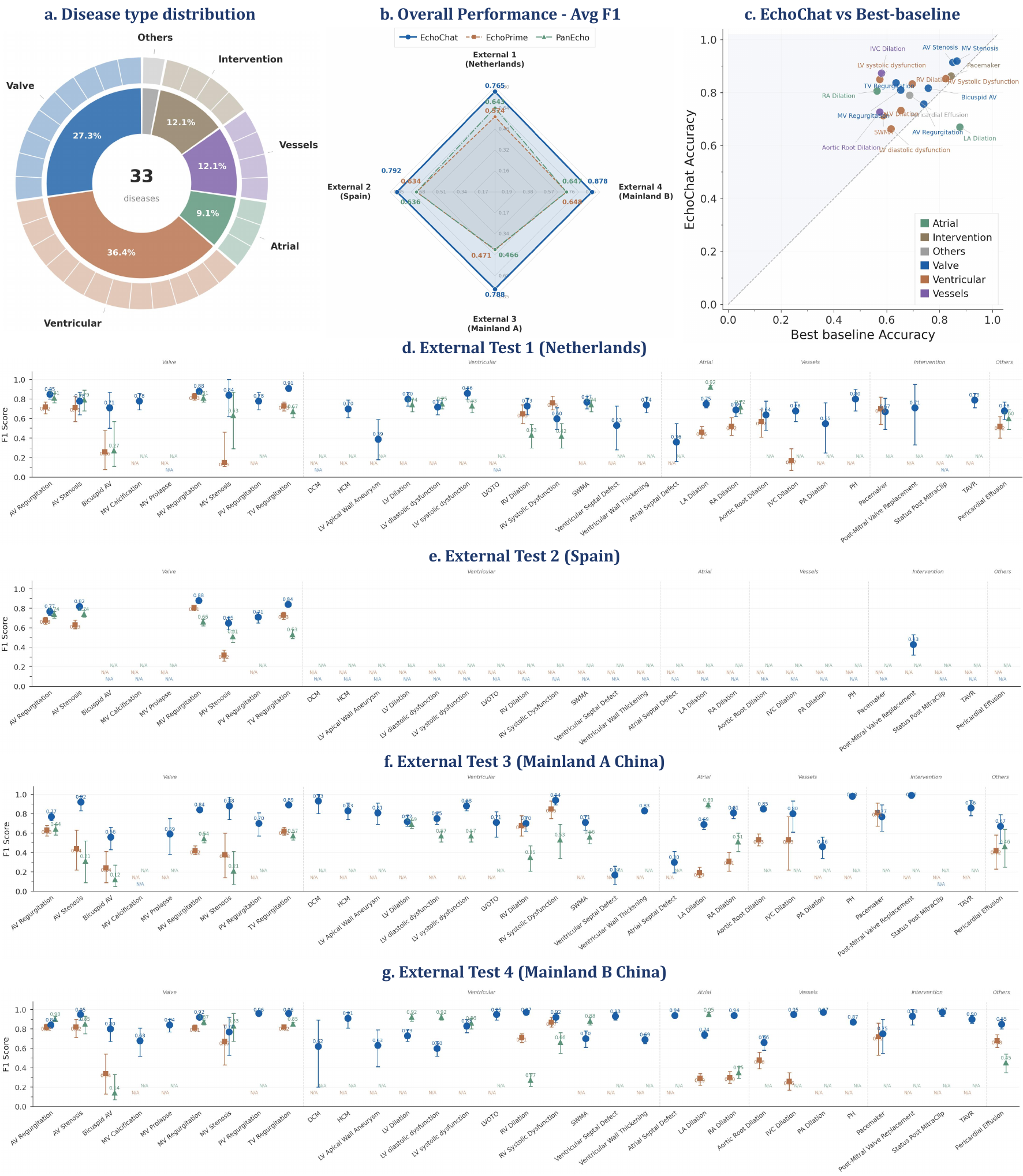}
    \caption{\textbf{Cardiac condition recognition task performance of \methodname{} and baseline methods.}\textbf{ a,} Distribution of the 33 cardiac conditions supported by \methodname{}, grouped into six clinical categories.
    \textbf{b,} Mean F1 scores of \methodname{} and baseline models in the four external test sets (the Netherlands, Spain, Mainland A China and Mainland B China), using the common subset of available conditions supported by all three models within each center.
    \textbf{c,} Condition-level accuracy comparison for conditions with an available baseline result. Each point represents one condition, with accuracy averaged equally across available external centers. The best-performing baseline is selected separately for each condition, and EchoChat and that baseline are compared on the same available centers. The x axis shows baseline accuracy and the y axis shows EchoChat accuracy; points above the diagonal favor EchoChat.
    \textbf{d-g,} Condition-level F1 scores of \methodname{} and baseline models on four external test sets.
    IVC: Inferior Vena Cava, RV: Right Ventricle, RA: Right Atrial, LA: Left Atrial, LV: Left Ventricle, MV: Mitral Valve, TV: Tricuspid Valve, AV: Aortic Valve, PV: Pulmonary Valve, PA: Pulmonary Artery, HCM: Hypertrophic Cardiomyopathy, TAVR: Transcatheter Aortic Valve Replacement, SWMA: Segmental Wall Motion Abnormality, PH: Pulmonary Hypertension, VSD: Ventricular Septal Defect, PE: Pericardial Effusion. NA: Not Available.}
    \label{fig:disease}

\end{figure}

As shown in~\Cref{fig:disease}, \methodname{} demonstrated cardiac condition recognition capabilities across diverse pathologies in the internal held-out Hong Kong SAR China cohort. The model showed particular strength in identifying valvular diseases, achieving precision/recall of 0.98/0.87 for aortic regurgitation, 0.97/0.98 for mitral regurgitation and 0.96/0.99 for tricuspid regurgitation. It also identified aortic stenosis and mitral stenosis with precision/recall of 0.96/0.97 and 1.00/0.98, respectively. Structural assessments included right atrial and left ventricular dilation, with precision/recall of 0.99/0.87 and 0.98/0.94. For post-intervention status, \methodname{} achieved precision/recall of 0.97/0.96 for transcatheter aortic valve replacement (TAVR) and 1.00/0.98 for mitral valve replacement.

Notably, \methodname{} achieved higher average F1 scores than current echocardiography foundation models on shared condition-recognition tasks across four external centers: the Netherlands, Spain, Mainland A China and Mainland B China. Across 50 condition-center entries shared with PanEcho~\cite{holste2025complete}, equally weighted mean F1 was 79.6\% versus 62.9\%; across 47 entries shared with EchoPrime~\cite{vukadinovic2026comprehensive}, it was 79.7\% versus 56.5\%, corresponding to absolute gains of 16.7\% and 23.2\%, respectively. For mitral regurgitation, \methodname{} achieved F1 scores of 0.88, 0.88, 0.84 and 0.92 in these four cohorts, compared with 0.81, 0.66, 0.54 and 0.87 for PanEcho and 0.83, 0.81, 0.42 and 0.81 for EchoPrime.
For right ventricular dilation, absolute F1 gains were 30\%, 35\% and 70\% over PanEcho and 8\%, 2\% and 26\% over EchoPrime in the Netherlands, Mainland A China and Mainland B China, respectively.
Beyond the shared tasks evaluated against PanEcho~\cite{holste2025complete} and EchoPrime~\cite{vukadinovic2026comprehensive}, we assessed EchoChat on 15 further conditions, including cardiomyopathies, congenital septal defects and post-intervention states.
Across 41 available external condition-center evaluations for these tasks, the mean F1 was 73.7\%.
Together, these results demonstrate improved average recognition on shared tasks and extend the evaluation to a broader range of clinically relevant findings (\Cref{fig:disease}).

Beyond binary condition recognition, \methodname{} graded the severity of 11 cardiac abnormalities using condition-specific ordinal scales. At the $\geq$mild and $\geq$moderate thresholds, internal-test F1 scores were 0.86/0.86 for left ventricular systolic dysfunction, 0.96/0.73 for mitral regurgitation and 0.64/0.78 for aortic stenosis. At the $\geq$moderate threshold, the corresponding precision/recall values were 0.90/0.82, 0.62/0.87 and 0.69/0.90. Four-grade exact agreement was 81.8\%, 85.3\% and 93.2\%, respectively; agreement within one grade was 99.2\%, 99.8\% and 99.3\%, with errors of at least two grades in 0.8\%, 0.2\% and 0.7\% of assessments. Linear-weighted Cohen's $\kappa$ was 0.775, 0.681 and 0.680, respectively. These results extend study-level interpretation to severity characterization.

EchoChat supports quantitative measurement of 22 echocardiographic parameters spanning key components of routine clinical reporting, including cardiac chamber size and wall thickness, left and right ventricular systolic function, and Doppler indices of diastolic filling and valvular and outflow-tract hemodynamics. In the internal Hong Kong SAR China test set, left ventricular ejection fraction (LVEF) estimation achieved a mean absolute error (MAE) of 2.11 (95\% CI, 1.94-2.28) and a coefficient of determination ($R^2$) of 0.90 (95\% CI, 0.88--0.91). Doppler-derived hemodynamic parameters were estimated with low error: mitral E-wave peak velocity had an MAE of 0.03~m/s and an $R^2$ of 0.94, while tricuspid regurgitation peak pressure gradient had an MAE of 1.53~mmHg and an $R^2$ of 0.93. For structural assessments, left atrial anteroposterior diameter had an MAE of 0.26~cm and an $R^2$ of 0.83, and interventricular septal thickness had an MAE of 0.14~cm. These results demonstrate quantitative assessment of cardiac function, structure and blood flow within a unified model.

\begin{figure}[!htbp]

    \centering
    \includegraphics[width=\textwidth,height=0.70\textheight,keepaspectratio]{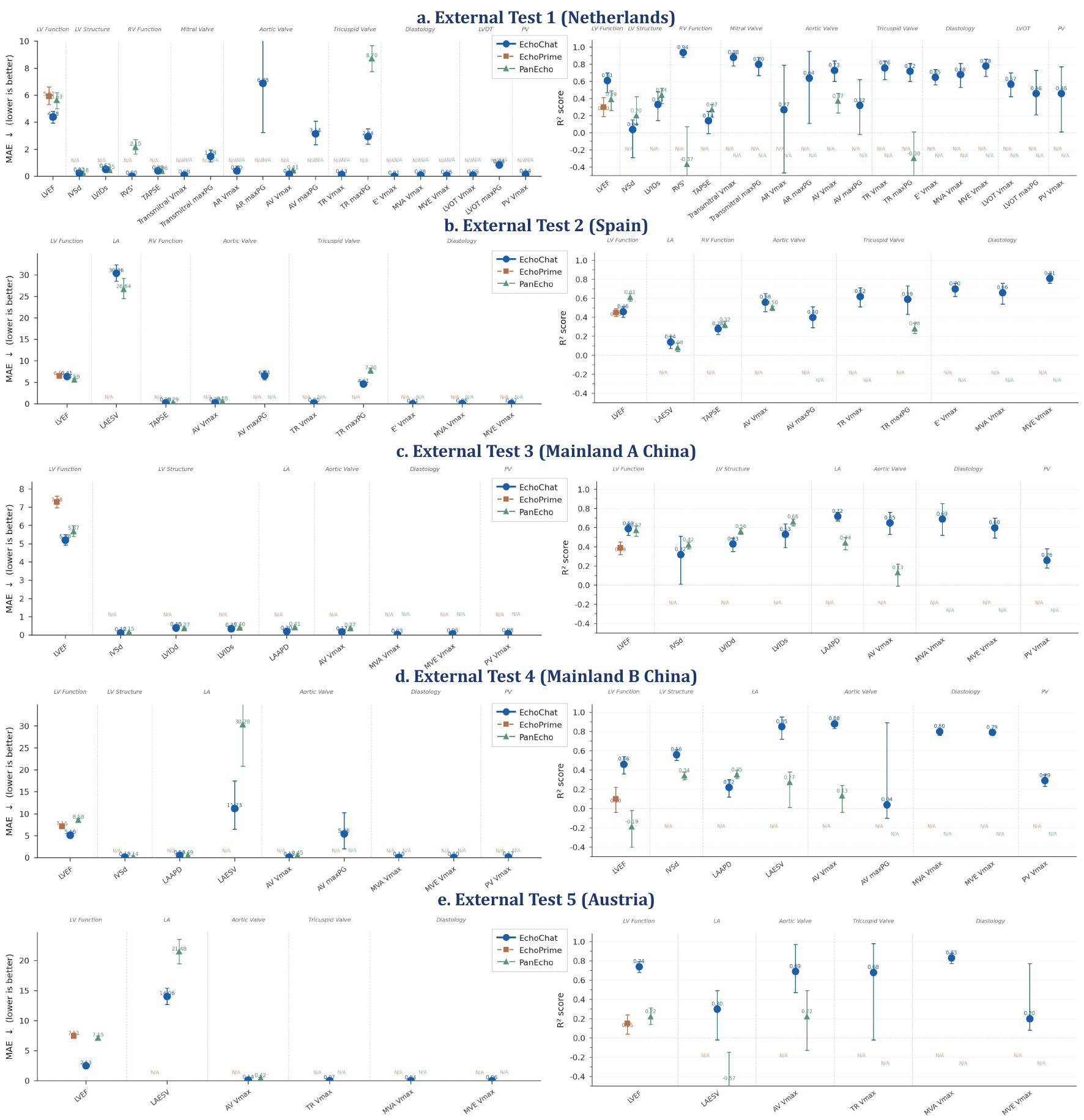}
    \caption{\textbf{Heart measurement estimation performance comparison of \methodname{} against baselines using Mean Absolute Error (MAE) and coefficient of determination ($R^2$).} \textbf{a-e,} Performance comparison on five external test sets from the Netherlands (\textbf{a}), Spain (\textbf{b}), Mainland China A (\textbf{c}), Mainland China B (\textbf{d}), and Austria (\textbf{e}). For each test set, the left panel shows the MAE (lower is better), while the right panel shows the $R^2$ score (higher is better). Error bars represent the 95\% confidence interval for each data point in the metrics. RV: Right Ventricle, RA: Right Atrial, LA: Left Atrial, LV: Left Ventricle, MV: Mitral Valve, AV: Aortic Valve, AAO: Ascending Aorta, PA: Pulmonary Artery, PV: Pulmonary Valve, LVEDD: Left Ventricular End-Diastolic Diameter, LVOT: Left Ventricular Outflow Tract, TV: Tricuspid Valve, TR: Tricuspid Regurgitation, PR: Pulmonary Regurgitation, AR: Aortic Regurgitation, IVS: Interventricular Septum, LVPWd: Left Ventricular Posterior Wall diameter at end-diastole, LVEF: Left Ventricular Ejection Fraction, LAAPD: Left Atrial Anteroposterior Diameter, TAPSE: Tricuspid Annular Plane Systolic Excursion, LAESV: LA end-systolic volume. NA: Not Available.}
    \label{fig:measurement}

\end{figure}

Compared with previous foundation models, \methodname{} achieved higher average measurement estimation performance on shared tasks across five external clinical centers (\Cref{fig:measurement}). Equally weighting all available shared measurement-center entries, mean $R^2$ was 0.518 for \methodname{} versus 0.244 for PanEcho across 26 entries, and 0.572 versus 0.278 for EchoPrime across the five LVEF entries supported in this evaluation. In Doppler assessment, \methodname{} estimated aortic valve peak velocity with MAEs of 0.15~m/s and 0.29~m/s in the Netherlands and Spain, compared with 0.41~m/s and 0.55~m/s for PanEcho. Tricuspid regurgitation peak pressure gradient also had lower MAEs in these cohorts (2.94 versus 8.70~mmHg and 4.61 versus 7.70~mmHg, respectively). In Mainland B China and Austria, left atrial end-systolic volume MAEs were 11.23 and 14.06~ml, compared with 30.28 and 21.48~ml for PanEcho. Together, these comparisons support improved average quantification across the shared measurements.

\bn{For left ventricular ejection fraction (LVEF) estimation, \methodname{} generalized across geographically diverse external cohorts, achieving the lowest MAE and highest coefficient of determination ($R^2$) among the evaluated foundation models in four of the five external clinical cohorts. Across these cohorts, MAE ranged from 2.53\% to 6.41\%, with $R^2$ values between 0.46 and 0.74. The advantage was particularly pronounced in Austria, where \methodname{} achieved an MAE of 2.53\% and an $R^2$ of 0.74, compared with 7.51/0.15 for EchoPrime and 7.15/0.22 for PanEcho. In Mainland B China, \methodname{} achieved an $R^2$ of 0.46, compared with 0.10 for EchoPrime and $-0.19$ for PanEcho. Scatter plots showed that predictions broadly tracked reference LVEF across reduced and preserved systolic function, with greater dispersion and compression at higher reference values in some external cohorts. Bland-Altman analysis revealed external mean biases ranging from $-4.11$ to $+1.20$ pp, with underestimation in the two mainland Chinese cohorts. Spain showed near-zero mean bias ($+0.44$ pp) alongside the widest 95\% limits of agreement ($-16.28$ to $+17.15$ pp), illustrating that small average bias can coexist with substantial individual differences. On the public EchoNet-Dynamic benchmark, \methodname{} achieved an MAE of 5.94\% and an $R^2$ of 0.58, outperforming EchoPrime (6.51/0.51) and remaining competitive with PanEcho (5.53/0.60), further supporting its applicability to single-video LVEF estimation.}

\subsection{\methodname{} supports structured report drafting and case-specific interaction with echocardiographers}
\label{sec:rgvqa}

\methodname{} generates structured reports from complete echocardiography studies with substantially higher abnormality-finding accuracy and completeness than Qwen2.5-VL~\cite{bai2025qwen2} on the internal held-out test set from Hong Kong SAR, China. Using finding-level matching to reduce the influence of reporting style, \methodname{} achieves an abnormality-finding F1 of 0.7341 (95\% CI, 0.7235-0.7448), compared with 0.1189 (95\% CI, 0.1080-0.1305) for Qwen2.5-VL. This improvement reflects gains in both abnormality-finding precision (0.7678 versus 0.2028) and recall (0.7387 versus 0.1093), indicating stronger agreement with reference findings and more complete coverage of report content. \methodname{} also outperforms Qwen2.5-VL across all four NLG metrics, achieving a BLEU-1 score of 0.7335 versus 0.1726, a 56.09\% absolute improvement, alongside higher ROUGE-1, ROUGE-L, and METEOR scores.

The case studies illustrate how reports generated by \methodname{} capture clinically relevant findings within a standard reporting structure across diverse cardiac conditions. In case~1, the generated report describes posterior mitral leaflet prolapse with severe mitral regurgitation, moderate tricuspid regurgitation, and severe left atrial dilatation, and integrates the principal valvular abnormalities into its impression. In case~2, it captures normal ventricular size and function alongside mild mitral and tricuspid regurgitation, matching the reference left ventricular ejection fraction of 64\% and right ventricular systolic pressure of 20~mmHg. Cases~3 and~4 illustrate the preservation of key functional findings alongside specific discrepancies in their detailed characterization. In case~3, \methodname{} correctly identifies mild concentric left ventricular hypertrophy and Doppler evidence of impaired relaxation, with an ejection fraction of 62\% matching the reference, but omits aortic leaflet thickening and describes mitral regurgitation as trace rather than mild. In case~4, it captures reduced left ventricular systolic function, reporting an ejection fraction of 35\% compared with the reference value of 36\%, together with mild aortic, mitral, and tricuspid regurgitation. Discrepancies in this case concern the degree of ventricular dilatation (mild versus moderate) and the severity and distribution of wall motion abnormalities (moderate global versus severe apical hypokinesis). Together, these examples show that \methodname{} can report key cardiac findings and characterize morphological abnormalities in a structured format. They also point to areas for further refinement, including regional localization and fine-grained severity assessment across a wider range of cardiac abnormalities, and they underscore that clinician oversight remains essential.

\methodname{} supports conversational access to echocardiographic findings through visual question answering (VQA). We evaluated this capability on the internal test set across eight clinically relevant question categories: heart chambers, Doppler imaging, cardiac motion, pulmonary assessment, valves, aorta and vasculature, congenital abnormalities and shunts, and systemic veins and volume status. These categories encompass questions about anatomy, function and haemodynamics across echocardiographic modalities. \methodname{} outperformed Qwen2.5-VL across all eight categories on all six evaluation metrics, covering lexical overlap, semantic similarity and factual accuracy. In Doppler imaging, for example, \methodname{} achieved a BLEU-1 score of 0.752 (95\% CI, 0.727-0.773), compared with 0.241 (0.226-0.255) for Qwen2.5-VL, and a token-level F1 score of 0.829 (0.810-0.844), compared with 0.359 (0.346-0.372).

To assess agreement with the underlying report findings, we also measured factual accuracy using Qwen3.6-27B to judge the consistency of model responses with report-derived reference answers under a predefined rubric, with complete definition in `Evaluation metrics and statistics' section. \methodname{} achieved higher factual accuracy in every category, with absolute improvements of 25.30\%-75.79\% over Qwen2.5-VL. The largest gain occurred in Doppler imaging, where factual accuracy reached 85.05\% (95\% CI, 81.89-88.15\%), compared with 9.26\% (6.72-12.05\%) for the baseline. The highest factual accuracy was observed for systemic veins and volume status, at 94.98\% (92.31-97.32\%), compared with 30.43\% (25.42-35.79\%). These results demonstrate that \methodname{}'s improvements in language-based evaluation are accompanied by substantially stronger agreement with report-derived findings across diverse echocardiographic questions. We further evaluated \methodname{} on MIMIC-Echo-VQA~\cite{mimic-echo-vqa}, using direct inference on 622 held-out multiple-choice echocardiography video questions and comparison with published benchmark results under their reported inference settings. Using direct inference, \methodname{} achieved the highest accuracy among the evaluated models at 66.4\%. It beat the strongest competing method, the agentic system Echo-CoPilot~\cite{HeiMoe_EchoCoPilot_MICCAISAT2026} (54.0\%), by 12.4\% and the best domain-adapted model, Qwen2-VL-7B-Biomed~\cite{thapa2025well} (49.0\%), by 17.4\%. \methodname{} also substantially outperformed general-purpose proprietary and open-source models, including GPT-o4-mini~\cite{o4mini} (43.9\%), Kimi K3~\cite{team2026kimi} (43.7\%), and GPT-4o~\cite{hurst2024gpt} (41.6\%). It exceeded Qwen2.5-VL-7B-Instruct~\cite{bai2025qwen2} under direct inference (34.0\%) by 32.4 points.

In the example cases, the answers generated by \methodname{} are consistent with the ground truth, capturing both qualitative findings and reported measurements, such as a left ventricular ejection fraction of 54\%. Representative findings include impaired ventricular relaxation, apical akinesis, mild multivalvular regurgitation, and left-to-right shunt flow across a ventricular septal defect. The responses also connect anatomical observations to physiological implications, such as associating inferior vena cava dilation with increased right atrial pressure, illustrating clinically relevant interpretation across multiple assessment domains.

\subsection{\methodname{} generalizes across external hospitals in cardiac condition recognition and measurement assessment}

We further evaluated \methodname{}’s recognition of the 33 cardiac conditions and findings across four external cohorts in the Netherlands, Spain, Mainland A China and Mainland B China, covering 33 conditions across the available cohort-specific evaluations. Mitral regurgitation precision/recall was 0.95/0.82, 0.81/0.96, 0.90/0.79 and 0.97/0.88, respectively, while tricuspid regurgitation F1 ranged from 0.84 to 0.96. Aortic stenosis F1 was 0.78, 0.82, 0.92 and 0.95 in these cohorts. The broader diagnostic scope also included hypertrophic cardiomyopathy, with F1 scores of 0.70, 0.83 and 0.91 in the Netherlands, Mainland A China and Mainland B China, compared with 0.92 internally. Across the same three external cohorts, F1 scores were 0.53/0.17/0.93 for ventricular septal defect, 0.36/0.30/0.94 for atrial septal defect and 0.39/0.81/0.63 for left ventricular apical aneurysm, characterizing the condition-specific variation across centers. Post-intervention recognition included TAVR, with F1 scores of 0.79, 0.86 and 0.90, respectively (\Cref{fig:disease}).

Severity assessment extended to the Netherlands, Mainland A China and Mainland B China cohorts. At the $\geq$moderate threshold, left ventricular systolic dysfunction F1 scores were 0.60, 0.72 and 0.61, compared with 0.49, 0.48 and 0.39 for PanEcho. For mitral regurgitation, F1 at the $\geq$mild threshold was 0.85, 0.84 and 0.87, compared with 0.65, 0.47 and 0.64 for PanEcho; at the $\geq$moderate threshold, \methodname{} achieved F1 scores of 0.48, 0.63 and 0.71. Four-grade mitral regurgitation assessment achieved exact agreement of 80.0\%, 77.9\% and 76.3\%, agreement within one grade of 98.6\%, 98.2\% and 95.4\%, and errors of at least two grades in 1.4\%, 1.8\% and 4.6\% of assessments, respectively. Corresponding linear-weighted $\kappa$ values were 0.669, 0.645 and 0.689. For aortic stenosis, $\geq$moderate F1 scores were 0.45, 0.36 and 0.44 using EchoChat's original four-grade outputs. For ordinal agreement, both models were evaluated on the shared three-grade scale of none/normal, mild/moderate and severe; EchoChat's linear-weighted $\kappa$ values were 0.255, 0.364 and 0.341.

We also evaluated \methodname{}’s estimation of the 22 echocardiographic parameters across five external centers, adding Austria and assessing each parameter where reference measurements were available. For LVEF, MAE/$R^2$ values were 4.38~pp/0.61 in the Netherlands, 6.41~pp/0.46 in Spain, 5.20~pp/0.59 in Mainland A China, 5.16~pp/0.46 in Mainland B China and 2.53~pp/0.74 in Austria. Hemodynamic measurements further demonstrated the breadth of external validation: mitral E-wave peak velocity had MAEs of 0.06-0.10~m/s across these five centers, while mitral A-wave peak velocity had MAEs of 0.03-0.10~m/s and $R^2$ values of 0.66-0.83. Aortic valve peak velocity achieved MAEs of 0.14-0.29~m/s and $R^2$ values of 0.56-0.88 across the five cohorts (\Cref{fig:measurement}).

\subsection{\methodname{} combines domain knowledge and structured reasoning to improve heart condition recognition and support diagnostic review}

The Echocardiography Knowledge Library (EKL) and Cardiologist-Like Reasoning (CLR) contributed complementary gains in heart condition recognition. Compared with the base model without both components in the training corpus, the EKL-only configuration achieved higher recall and F1 ($p_{\mathrm{Holm}}=0.0012$ and $0.0160$, respectively), whereas the CLR-only configuration achieved higher accuracy, recall and F1 ($p_{\mathrm{Holm}}=0.0012$, $0.0114$ and $0.0042$, respectively). These comparisons used Holm correction across 12 tests against the base model. Adding either component when the other was already present further improved accuracy, recall and F1 (all $p_{\mathrm{Holm}}=0.0016$ under a separate 16-test correction). The combined model achieved the highest accuracy (87.56\%), recall (82.86\%) and F1 (87.13\%) among the evaluated configurations, with significant improvements over the base model in all three metrics (all $p_{\mathrm{Holm}}=0.0012$). These gains support the joint contribution of domain knowledge and structured reasoning to diagnostic performance.

CLR also made the stated findings and diagnostic inferences available for inspection. In three illustrative cases, EchoChat linked an ejection fraction of 17\% and severe global hypokinesis to left ventricular systolic dysfunction, identified reference-reported mild aortic stenosis while citing a mean gradient of 12~mmHg, and recognized a small pericardial effusion while discussing its distinction from tamponade. All three final answers agreed with the reference answers, and the accompanying rationales articulated the findings and clinical interpretations supporting each conclusion.

Cases with incorrect predictions demonstrated how these explanations support error analysis. EchoChat excluded atrial fibrillation based on normal atrial size and an asserted absence of rhythm abnormalities, despite reference-documented atrial fibrillation at 120~beats per minute. It also asserted mild concentric left ventricular hypertrophy despite reference-reported normal wall thickness and denied pericardial effusion despite a documented thin rim of fluid. By exposing unsupported diagnostic exclusions and discrepancies in the stated imaging findings, CLR enables targeted review of the rationale accompanying each prediction.

\subsection{\methodname{}-echocardiographer collaboration improves cardiac condition recognition}

\begin{figure}[!htbp]

    \centering
    \includegraphics[width=\textwidth,height=0.70\textheight,keepaspectratio]{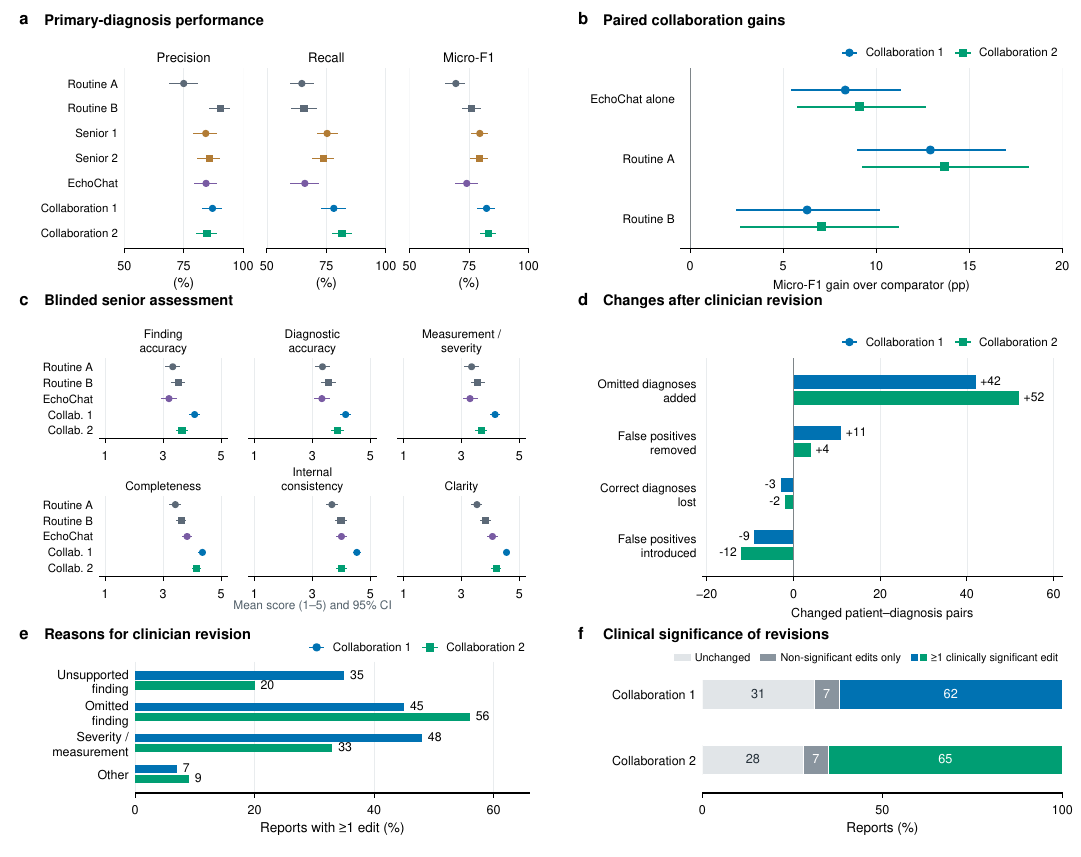}
    \caption{\textbf{Echocardiographer-EchoChat collaboration and individual echocardiographer benchmarks.}
\textbf{a,} Precision, recall and micro-F1 for two routine echocardiographers, two senior echocardiographers, EchoChat, and two routine echocardiographer-EchoChat collaborations on 100 examinations. All conditions use the same 21 clinician-defined primary diagnostic targets. Symbols identify individual outputs; horizontal lines show patient-bootstrap 95\% confidence intervals. All three axes span 50-100\%.
\textbf{b,} Patient-paired micro-F1 gains for each collaboration over the comparator named on the y axis: EchoChat alone, routine echocardiographer A, or routine echocardiographer B. Positive differences favour collaboration. Lines show 95\% patient-bootstrap confidence intervals; pp denotes percentage points.
\textbf{c,} Source-blinded senior assessment based on the six-dimension rubrics. Means and patient-bootstrap intervals are shown separately by dimension. Each point averages both raters within patient and then across patients; error bars show 95\% patient-bootstrap confidence intervals. Scores range from 1 to 5, with higher values indicating better quality.
\textbf{d,} Label changes from the initial EchoChat draft to each final collaborative report, evaluated against the primary-diagnosis reference labels. Omitted diagnoses added are FN-to-TP transitions; false positives removed are FP-to-TN; correct diagnoses lost are TP-to-FN; false positives introduced are TN-to-FP. Rightward positive bars indicate corrections and leftward negative bars indicate new disagreements.
\textbf{e,} Reports with at least one edit of each clinician-recorded reason. Denominators are 100 saved reports per collaborator; categories overlap, so percentages need not sum to 100. Counts refer to reports, not editing operations.
\textbf{f,} All reports partitioned into unchanged, only edits marked non-significant, or at least one edit marked clinically significant by the revising clinician. Numbers inside bars are counts out of 100.
    }
    \label{fig:userstudy}

\end{figure}

To evaluate \methodname{}'s assistive role, we compared seven outputs on the same held-out Mainland A China cohort of 100 echocardiographic examinations, excluded from Chinese report-adapter training: two standalone routine-reader outputs (Routine A/B), two standalone senior-reader outputs, EchoChat alone and two routine echocardiographer-EchoChat collaborations. All outputs were evaluated against the same 21 clinician-defined primary diagnostic labels, comprising 320 reference-positive patient-diagnosis pairs. The two collaborative outputs achieved micro-F1 scores of 82.4\% and 83.1\%, compared with 69.5\% and 76.1\% for Routine A/B, 79.5\% and 79.3\% for the senior readers, and 74.0\% for EchoChat alone (\Cref{fig:userstudy}a,b). Collaborative recall reached 78.1\% and 81.6\%, compared with 64.7\% and 65.6\% for Routine A/B, while precision was 87.1\% and 84.7\%. These results support clinician-reviewed AI drafts as a means of improving the coverage of primary diagnoses.

To characterize the computational requirements for this assistive workflow, we measured model inference on one NVIDIA H800 80~GB GPU with batch size one. A cardiac condition-recognition query took $1.939\pm1.869$~s (mean $\pm$ standard deviation; 300 queries), and generation of a complete ten-section report took $7.884\pm0.959$~s (500 queries), excluding preprocessing. Inputs comprised one 16-frame video for each of 11 main echocardiographic views and all static images in the study. These measurements quantify the model-inference component of producing an interpretation or draft report for clinician review.

\begin{figure}[!htbp]

    \centering
    \includegraphics[width=\textwidth,height=0.70\textheight,keepaspectratio]{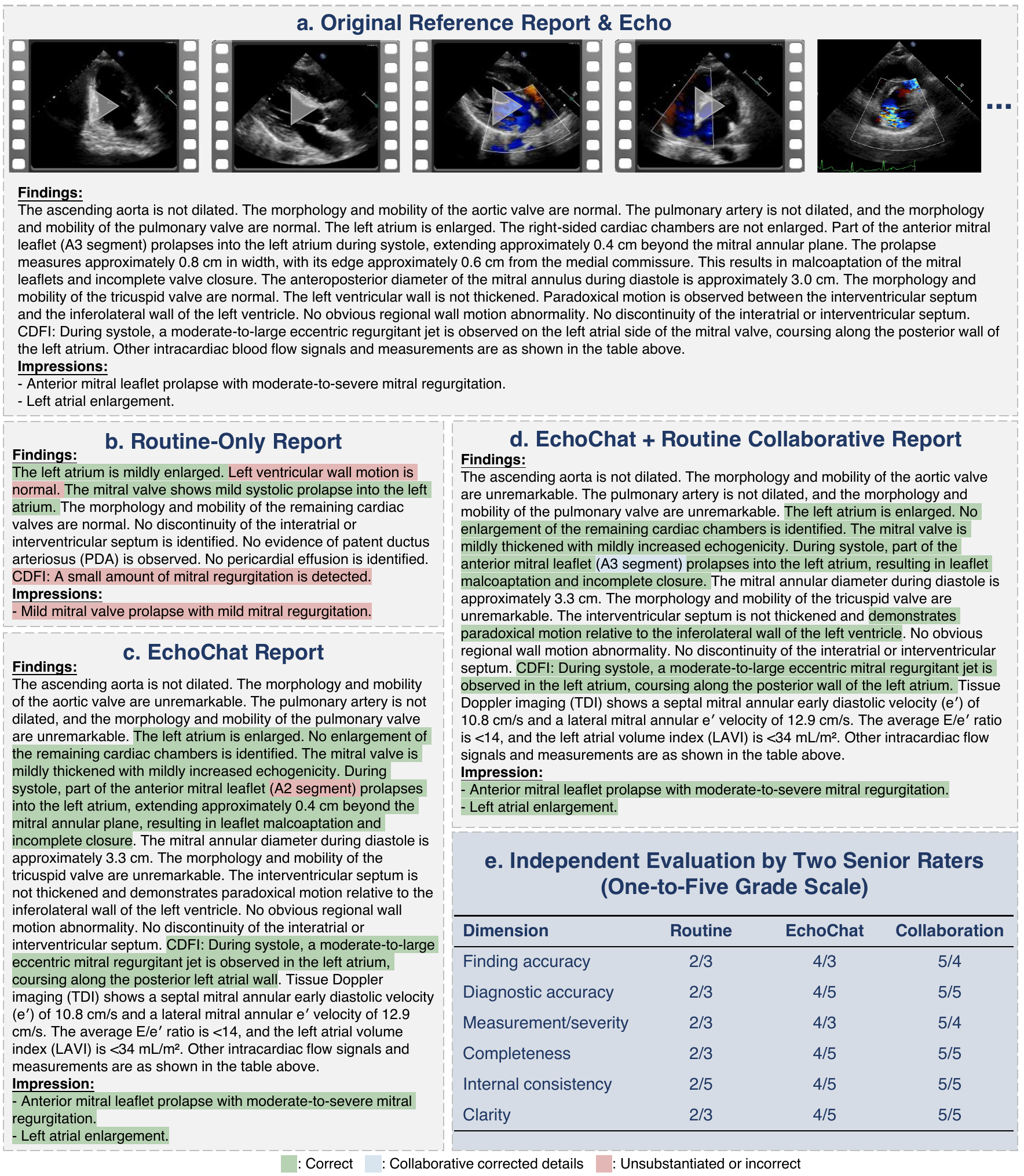}
    \caption{\textbf{Representative case illustrating EchoChat-routine echocardiographer collaboration.}
    a, Representative echocardiographic input images, and clinical reference report from PACS. b, Standalone report from the routine echocardiographer. c, Report generated by \methodname{} from the echocardiographic examination. d, Collaboration report, produced by a routine echocardiographer revising the EchoChat draft. e, Independent assessment by senior raters across finding accuracy, diagnostic accuracy, measurement/severity accuracy, completeness, internal consistency, and clarity/report expression. Each cell reports the scores of the two senior raters.}
    \label{fig:user_study_exp}

\end{figure}

\subsection{\methodname{}-echocardiographer collaboration enhances echocardiography report quality}

\methodname{}'s clinical utility was further assessed through collaborative report revision and independent evaluation of report content in the same 100-examination cohort. Two routine echocardiographers revised EchoChat drafts to produce two collaborative reports per examination. Two source-blinded senior raters independently assessed five reports per examination: two standalone routine reports, the EchoChat draft and two collaborative reports, yielding 1,000 report assessments. Each assessment comprised six separate scores from 1 to 5: imaging finding accuracy, diagnostic conclusion accuracy, measurement/severity accuracy, completeness, internal consistency and clarity/report expression. Reports were presented in the same style and format, and raters had access to the original echocardiographic examinations.

The source-blinded quality assessment supported these gains in clinical content.
Averaged across the two outputs in each group, mean diagnostic-accuracy scores were 15.84\% higher for collaboration than for standalone routine reporting: 4.15 and 3.86 for Collaborations 1/2, compared with 3.35 and 3.57 for Routine A/B.
All four paired comparisons favored collaboration, with gains of 0.30-0.80 points (Holm-adjusted $P\leq0.0424$);
the largest gain was 23.88\% (0.80 points; 95\% CI, 0.59-1.02) for Collaboration 1 versus Routine A.
Mean completeness scores were 20.74\% higher for collaboration than for standalone routine reporting: 4.36 and 4.15 versus 3.42 and 3.62, with gains of 0.53-0.94 points (all four adjusted $P=0.0024$). Scores were averaged across the two raters within each examination before patient-paired comparisons (\Cref{fig:userstudy}c).

The remaining quality dimensions characterized distinct aspects of collaborative reporting.
Mean finding-accuracy scores increased by 12.92\% relative to standalone routine reports, with gains ranging from 0.13 to 0.76 points and significant differences in three of the four adjusted comparisons;
Collaboration 2 versus Routine B had a gain of 3.69\% (0.13 points; 95\% CI, $-0.08$ to 0.36; adjusted $P=0.2806$).
Mean measurement/severity-accuracy scores increased by 13.38\% relative to standalone routine reports, with gains ranging from 0.12 to 0.81 points and three comparisons significant after Holm correction (adjusted $P\leq0.0275$).
Mean internal-consistency scores increased by 11.55\% relative to standalone routine reports: 4.54 and 4.02 for Collaborations 1/2 versus 3.68 and 3.99 for Routine A/B, with significant gains in three comparisons.
Clarity/report expression was assessed separately and showed higher mean scores for both collaborative outputs than for either routine output (\Cref{fig:userstudy}c).

Clinician revision improved diagnostic coverage by adding reference-supported diagnoses to the initial EchoChat drafts (\Cref{fig:userstudy}d). Collaborations 1/2 added 42/52 omitted diagnoses and removed 11/4 false positives, while losing 3/2 correct diagnoses and introducing 9/12 false positives. These changes count patient-diagnosis pairs across the 21 primary labels. They reconcile the initial EchoChat counts of 211 true positives, 39 false positives and 109 false negatives with 250/37/70 for Collaboration 1 and 261/47/59 for Collaboration 2.

Clinician-recorded editing reasons further showed how the drafts were refined (\Cref{fig:userstudy}e). Among 100 saved reports per collaborator, Collaborations 1/2 included edits addressing unsupported findings in 35/20 reports, omitted findings in 45/56 reports, severity or measurement discrepancies in 48/33 reports and other reasons in 7/9 reports. These categories overlap because a report can contain more than one reason for revision. Separately, the revising clinicians classified 62/65 reports as containing at least one clinically significant edit, 7/7 as containing only non-significant edits and 31/28 as unchanged (\Cref{fig:userstudy}f). These report-level classifications complement the reference-based diagnosis changes and the independent senior assessments of clinical content.

A representative case compares the routine echocardiographer-only report, the EchoChat draft and the EchoChat-routine echocardiographer collaboration (\Cref{fig:user_study_exp}), illustrating complementary contributions to the collaborative report. The reference described A3 anterior mitral-leaflet prolapse with moderate-to-severe mitral regurgitation, whereas the selected routine-only report described slight prolapse and mild regurgitation. The initial EchoChat draft described moderate-to-severe regurgitation but localized the prolapse to A2. During collaboration, the echocardiographer corrected A2 to A3 while retaining the regurgitation severity, bringing both findings into agreement with the reference. Across the two senior raters, mean finding-accuracy and measurement/severity-accuracy scores were each 4.5/5 for the collaborative report, 3.5/5 for the AI draft and 2.5/5 for the selected routine-only report. The case demonstrates how clinician revision can preserve useful AI content while refining anatomical localization.

\section{Discussion}

\methodname{} is an echocardiography Multi-modal Large Language Model (MLLM) capable of analyzing a complete echocardiography study, which consists of diverse cardiac views and examination modalities, by leveraging a foundation model with a large parameter scale. \bn{Trained on 46,358 echocardiographic examinations within a collection of 58,768 examinations from six centers}, \methodname{} extends the evaluated scope of previous echocardiography foundation models while achieving higher average performance on shared recognition and measurement tasks~\cite{vukadinovic2026comprehensive,holste2025complete}. Moreover, \methodname{} introduces the novel capability to generate comprehensive echocardiography reports and answers questions about echocardiographic findings, extending the scope of echocardiographic AI assistance. \bn{Results from evaluations at five international external test centers, with condition recognition assessed at four and measurements at all five, support \methodname{}'s generalizability and robustness across different demographic groups and regional shifts. In the \methodname{}-routine echocardiographer collaboration reader study, primary diagnosis conclusion from collaborative reports achieved an averaged micro-F1 score of 82.75\%, having statistically significant 9.97 percentage-point gain compared with routine-reader only reports (72.78\%). Two source-blinded senior raters also found higher diagnostic accuracy and completeness in collaborative reports (\Cref{fig:userstudy}c).} Together, these results support \methodname{} as an assistant for clinician-reviewed interpretation and report preparation.

A key bottleneck in transitioning previous echocardiography foundation models to clinical deployment is their insufficient support for cardiac conditions and heart measurements. Previous echocardiography foundation models supported up to 18 conditions and 21 measurements~\cite{vukadinovic2026comprehensive,holste2025complete}, while task-specific models addressed individual problems such as aortic, mitral and tricuspid regurgitation~\cite{binder2025automated,vrudhula2024high,vrudhula2025automated} or left ventricular ejection fraction estimation~\cite{ouyang2020video}. Meanwhile, echocardiographic clinical interpretation draws on a wider combination of structural, functional and hemodynamic findings~\cite{mitchell2019guidelines,robinson2020practical,houska20242024}.
The limited scope and robustness of current echocardiography AI in supporting cardiac conditions and measurements present a critical gap between these technologies and clinical practice.
\bn{By learning from large-scale echocardiography records, EchoChat integrates recognition of 33 cardiac conditions and findings, measurement of 22 echocardiographic parameters and severity grading for 11 abnormalities. Together, these capabilities span common valvular and ventricular abnormalities, congenital defects, cardiomyopathies and post-intervention states, alongside quantitative assessment of cardiac structure, function and blood flow relevant to routine clinical reporting. Evaluation beyond the shared baseline comparisons included 15 additional condition-recognition tasks, with a mean external F1 of 73.7\%. External validation also demonstrates higher average performance on shared tasks than both evaluated foundation models. }\methodname{} represents a significant step toward adopting a unified solution for the echocardiography diagnostic workflow, accommodating a broader range of patient scenarios.

Moreover, \methodname{} explores a new learning paradigm that is more aligned with the learning process of the clinician to better enhance the biomedical foundation model performance. Current biomedical foundation models are often trained on data collected from healthcare providers for diagnostic purposes, which often contain the input images and the final conclusions or labels~\cite{li2023llava,zhang2024generalist}, lacking crucial background information, such as the correlation between echocardiography findings and disease severity.
This training data constitution revealed three key limitations. First, the lack of background information misaligns the model's learning and decision process with that of clinicians~\cite{mitchell2019guidelines}, leading to a lack of essential medical knowledge required for interpreting multi-modal inputs and potentially unreliable predictions. Second, the label-only data, even with rigorous diagnostic procedures, is still prone to human-induced systematic annotation error~\cite{radsch2023labelling,sylolypavan2023impact,alber2025medical}, especially in large-scale training data, where high-quality annotation is hard to obtain. Third, only the model prediction is not sufficient for clinicians to make clinical decisions; in contrast, clinicians need evidence provided by the model to further adapt the model's output as a valid suggestion.
\bn{By incorporating a comprehensive echocardiography knowledge library (EKL), curated from authoritative sources, and echocardiography-based reasoning data into the training corpus, \methodname{} gains accurate, clean domain expertise with a lower risk of annotation errors. It also produces explicit reasoning trajectories that make its stated findings and inferences available for inspection. In the ablation analysis, combining these two components increased the F1 score from 81.63\% to 87.13\%.}

\bn{\methodname{} is designed to serve as an assistant to echocardiographers. It targets echocardiography, the most widely used cardiovascular imaging modality, which remains underrepresented in existing biomedical MLLMs~\cite{wang2025interpretable,yang2025multi,zhang2024generalist,li2023llava,singhal2023large,kim2024transparent}. Given a multiview, multimodal transthoracic echocardiography study, \methodname{} flags potential abnormalities, assesses severity grades, estimates cardiac measurements and drafts a structured report. Its visual question-answering capability lets echocardiographers ask about the study and look more closely at specific findings. Together, these capabilities support a workflow in which the echocardiographer reviews the underlying images, verifies the model's findings and measurements, revises the draft and finalizes the report. Clinical judgement and responsibility for the final report stay with the clinician throughout. Because \methodname{} helps with key steps of assessment and reporting, not only with recognizing individual cardiac conditions, it addresses a practical bottleneck. Interpreting echocardiographic studies efficiently and producing high-quality reports require substantial expertise and effort, particularly for complex cases~\cite{popescu2009european}, while demand for examinations and timely diagnoses remains high~\cite{ng2022feasibility}. The level of expertise involved is illustrated by recommendations that advanced (level III) training include the interpretation of 750 transthoracic echocardiography studies~\cite{wiegers20192019}. Consistent with this assistant role, reports prepared by echocardiographers working with \methodname{} improved statistically significantly on reports prepared without assistance.
Mean gains relative to standalone routine reports were 15.84\% (0.55 points) in diagnostic accuracy and 20.74\% (0.73 points) in completeness on five-point scales rated by senior echocardiographers (\Cref{fig:userstudy}c). These results suggest that \methodname{} can help echocardiographers produce more accurate and complete reports while leaving the final clinical decision in their hands.}

The results are not without limitations. \methodname{} requires standardized, high-resolution echocardiography records that are manually collected by clinicians to ensure high-quality echocardiography interpretations. Although it is clinically standard for sonographers to collect such high-quality records to support diagnoses with clear evidence, it remains valuable to explore the potential of MLLM-based echocardiography comprehension in diverse imaging settings and varying levels of quality.
\bn{The current user study represents a controlled simulation rather than real-world deployment in a live echocardiography laboratory. Further validation is needed to address important aspects of human-AI collaboration, such as automation bias, longitudinal model drift, integration with PACS and reporting systems, and patient outcomes. Multi-turn conversational evaluation is also a direction for future work.
Deployment may improve performance by prompting review of structures or abnormalities that might otherwise be overlooked and by standardizing both quantitative and descriptive reporting. However, performance may also be degraded by incomplete image acquisition and automation bias. Any clinical implementation should therefore incorporate assessment of image quality and view completeness, along with clinician verification before issuing the final report.}

\methodname{}'s results highlight a promising direction for further exploration of large-scale, expertise-infused multi-modal large language foundation models in cardiology. Large-scale multi-modality echocardiography studies provide rich cardiac information about health, while multi-modal large language models serve as high-throughput tools to efficiently mine correlations within echocardiography data and enhance comprehension. The integration of these two approaches has the potential to improve cardiovascular health understanding for individuals at all levels of cardiology expertise. Future work will aim to enhance the accuracy and robustness of MLLMs in interpreting comprehensive echocardiography studies, while exploring collaborative frameworks for integrating these models into cardiology departments to support echocardiographers in their diagnostic and interpretive workflows.

\section{Methods}

\subsection{Echocardiography Data Curation and Training Data Construction}

\noindent \textbf{Echocardiography raw data curation.}
Six centers across four countries participated in the study: two centers in mainland China, one in Hong Kong SAR China, one in Spain, one in Austria, and one in the Netherlands. Together, these centers contributed 58,768 examinations.
In Mainland China, Hong Kong SAR China, Netherlands, and Spain, each study consists of a standard, complete patient record that includes a final echocardiography report, a full heart parameter measurement record, and 30-150 echocardiographic videos and images. These videos and images visualize heart structures and motion patterns from various angles and locations using different imaging modalities, including 2D echocardiography, Color Doppler, Tissue Doppler, Pulsed Wave Doppler, and Continuous Wave Doppler. These comprehensive records support a full echocardiography-based diagnosis.
In Austria, each study consists of echocardiographic imaging paired with measurement records and contributes to measurement-only evaluation.
The data curation process includes all cardiac conditions recorded during clinical routines without further exclusion. The imaging and clinical records of each echocardiography study were utilized for the development and validation of \methodname{}.

\bn{Based on these inclusion criteria, we collected echocardiography data from two hospitals in Mainland China spanning 2019 to 2025, comprising 6,857 and 1,086 examinations with 200,808 and 33,572 echocardiographic videos and images, respectively.
We obtained 47,940 examinations with 2,190,296 echocardiographic videos and images at Hong Kong SAR China from 2016 to 2025.
From Netherlands, we collected 724 examinations with 49,768 echocardiographic videos and images from 2023.
From Spain, we collected 1,000 examinations with 50,158 echocardiographic videos and images spanning 2019 to 2023.
From Austria, we collected 1,161 examinations with 49,327 echocardiographic videos and images.}
This research is approved by the ethical committee of all participating hospitals. A waiver of consent was granted for the use of retrospective, de-identified data.

Data were assigned to base-model development, external evaluation and Chinese-adapter reader-study evaluation. The base model was trained on 46,358 examinations from Hong Kong SAR, China, with separate examinations from that center reserved for internal testing. Mainland A contributed 6,857 examinations in total: 100 were held out for the reader study, and the remaining 6,757 were used to train the Chinese report-generation adapter. The reader study used this adapter on the 100 held-out examinations without patient overlapping.
The five external test sets comprised 724 examinations from the Netherlands, 1,000 from Spain, 1,161 from Austria, 1,124 from Mainland A China and 1,086 from Mainland B China. The 1,124 Mainland A external-test examinations are a subset of that center's 6,857 examinations. External evaluation at every center, including Mainland A, used the base model without the Chinese report-generation adapter.

The model was distributed to individual external centers for local testing on their proprietary servers.
Examinations without detectable heart abnormalities were designated as the negative samples without any further screening for the evaluation dataset. The test sets for the severity stratification tasks are subsets of the full test sets, restricted to cases for which evidence supporting the ground-truth severity labels can be readily extracted.

All echocardiography data were exported from the respective hospitals' Picture Archiving and Communication System (PACS) in Digital Imaging and Communications in Medicine (DICOM) format. An automated processing pipeline was employed to convert these raw DICOM files into MPEG-4 (.mp4) and Portable Network Graphics (.png) formats. During this conversion process, ECG traces, and text information were masked to ensure privacy and prevent leakage of reported findings or measurements. The resulting echocardiographic videos and images were then resized into 224 $\times$ 224 RGB format to serve as input for the model. For videos, a total of 16 frames were sampled from consecutive 32 frames, using a sampling interval of 2. The video preprocessing strategy is consistent with the previous echocardiography foundation model~\cite{vukadinovic2026comprehensive}.
Measurements and echocardiography reports were also exported from the PACS systems in tabular format.
For the baseline result, we used their released original codes and followed their evaluation protocol to test on all external cohorts.

\bn{All cardiac condition labels are derived from the paired echocardiography full report.
For binary condition recognition, an explicitly documented abnormality in either the findings or summary is positive, including trace/trivial findings and abnormalities without a stated severity.
Severity labels were constructed for 11 abnormalities: aortic regurgitation, aortic stenosis, left atrial dilation, left ventricular dilation, left ventricular systolic dysfunction, mitral regurgitation, mitral valve calcification, pulmonary hypertension, right ventricular dilation, tricuspid regurgitation and ventricular wall thickening. Explicit descriptions were mapped to none/normal, mild, moderate or severe. Trace/trivial and mild-to-moderate descriptions were mapped to mild, and moderate-to-severe descriptions to moderate. Ungraded abnormalities were excluded from ordinal labels while retaining their positive binary labels. Left atrial dilation, left ventricular dilation, right ventricular dilation and ventricular wall thickening used three grades, combining moderate and severe as at least moderate. For external aortic stenosis ordinal agreement comparisons, both models were evaluated on the shared three-grade scale of none/normal, mild/moderate and severe following PanEcho~\cite{holste2025complete}. The mapped labels were used as targets for severity-grading prompts.
Note that for countries where the echocardiographic reports are not written in English, we first translate the full reports into English using Qwen3.6-27B~\cite{qwen3.6-27b} and then extract the cardiac condition labels using the same English report processing pipeline. Cardiologists at these external centers also provide mappings of cardiac measurement names from the local language to standardized English terminology. The resulting English cardiac condition labels and measurement names are verified by local cardiologists for each examination. This report-derived diagnostic-reference procedure applies to centers with reports; Austria provides measurement references.}
To ensure consistency across all centers, we standardized the diameter measurements to $cm$, velocity measurements to $m/s$, and area measurements to $cm^2$.

\noindent \textbf{\methodname{} training data construction.}
To construct the training data for \methodname{}, we further processed the exported echocardiography records. The Qwen3.6-27B~\cite{qwen3.6-27b} large language model processed the cleaned and anonymized full reports along with the provided prompt as input. The model then output standardized disease labels. These finalized labels were assigned to each echocardiography study for training data construction.
Additionally, we annotated each echocardiography video with view information using state-of-the-art echocardiography view recognition models~\cite{vukadinovic2026comprehensive,song2025echoviewclip}. When multiple videos were identified for the same view, the video with the highest view prediction confidence was selected as the model input. Model inputs comprise one selected video for each of the 11 main views, chosen by the view-confidence procedure described below, together with all static study images; the same input selection is used during training and inference.
These preprocessed records formed the basis for \methodname{}'s training data, designed to support six distinct clinical tasks:
Measurement Estimation: The model outputs specific heart parameter measurements in response to user inquiries.
Heart Condition Recognition: The model identifies the presence of specific heart conditions based on user queries.
\bn{Severity Grading: The model predicts the ordinal grade of a specified abnormality using the label definitions above.}
Report Generation: Given a complete echocardiography study, the model generates a comprehensive echocardiography report.
Echocardiography Visual Question Answering: The model answers user-specific echocardiography-related questions.
Diagnostic Reasoning: When asked about heart condition recognition, the model outlines the complete echocardiography-based decision-making process before providing the final diagnosis.
For Measurement Estimation, Heart Condition Recognition, Severity Grading, and Report Generation, we employed a template-based approach to construct the training data. To generate training samples, we first select the question and answer pairs specifically designed for each task. Next, we substituted the relevant item names and ground truth values, including mapped severity grades into the selected question-answer pair, creating a complete training instance.

To enhance the question diversity for Heart Condition Recognition and improve recognition performance, we incorporated multi-choice questions into the training data. Specifically, \bn{for each echocardiography study, we provided the Qwen3.6-27B~\cite{qwen3.6-27b} model with the prompt and the full corresponding echocardiography report.} The model then generated a set of answer candidates for multi-choice questions based on the standardized full-report labels. These candidates were used to replace placeholders in the question templates, resulting in diverse training data.

\bn{For Echocardiography Visual Question Answering, we designed eight categories of questions that echocardiographers are most likely to encounter during the diagnostic workflow: valve assessment, heart chambers assessment, heart motion assessment, pulmonary assessment, Doppler imaging assessment, aorta and vascular assessment, congenital abnormalities and shunts, and systemic veins and volume status.} These categories comprehensively cover the majority of heart structures and echocardiography imaging modalities, enabling the model to address a wide range of clinical inquiries.
To construct the training data, we utilized an LLM-based pipeline to transform clinical information from echocardiography reports into visual questions and corresponding ground truth answers. The Qwen3.6-27B model processed each echocardiography report along with the task-specific prompts. The model then generated structured open-ended question-answer pairs. After performing manual sampling to ensure the quality and accuracy of the outputs, the curated data were finalized as the training set for the Echocardiography Visual Question Answering task.

\subsection{Multi-Agent Collaborative Echocardiography Knowledge Injection}

\noindent\textbf{Constructing EKL.} We curated the largest-to-date Echocardiography Knowledge Library (EKL) to enhance model training by injecting high-quality domain knowledge alongside the large-scale patient data. This knowledge library is sourced from a diverse range of authoritative materials, including 14 textbooks, 91 clinical guidelines from the American Society of Echocardiography, European Society of Cardiology, and Chinese Medical Society, 573 echocardiography healthcare education entries from StatPearl, and 176 online case studies.
The EKL provides a comprehensive repository of information, covering over 100 cardiac conditions. These conditions range from commonly occurring issues, such as valve regurgitation, to rare congenital heart diseases. The library also encompasses more than five echocardiography modalities and more than thirty cardiac views, with content presented in various formats such as figures with captions, pure text, tables, and diagrams. Furthermore, it includes detailed knowledge required to perform and interpret echocardiographic examinations, including maneuver steps, imaging techniques, and diagnostic standards.

To systematically organize echocardiography knowledge in diverse formats, we developed a multi-agent collaborative curation pipeline. The pipeline takes raw knowledge sources as input and outputs a Unified Medical Language System (UMLS) concept-indexed database. This database is composed of entries with two key components: content and indexing concepts.
The content represents a segment of echocardiography knowledge and can take various forms, such as an image-caption pair, a pure-text paragraph, a table, or a diagram. The indexing concepts are a set of UMLS-standardized echocardiography-related terms that best describe the corresponding content. Users can easily retrieve the desired information from the database by using these indexing concepts as search terms.
Each agent in the pipeline is a state-of-the-art Multimodal Large Language Model (MLLM), powered by GPT-4o~\cite{hurst2024gpt} and Qwen 2.5-VL-72B~\cite{bai2025qwen2}. These agents are equipped with specially designed system prompts, which enhance their performance for specific curation tasks.

The workflow for constructing the Echocardiography Knowledge Library (EKL) involves a systematic four-step process to ensure the organization and quality of the curated knowledge. Initially, the raw content, including PDF textbooks and HTML files, is processed by the Content Extractor agent, which is powered by a state-of-the-art foundation model-based document processor~\cite{mistral}. This agent extracts pure text, image-caption pair candidates, tables, and diagrams from the source materials, forming the foundation for subsequent processing. Next, the Subfigure Identifier and Subcaption Splitter agents collaborate to address ambiguities caused by compound image-caption pairs, where multiple subfigures are combined into a single image. The Subcaption Splitter identifies potential subcaptions within the complete caption, which are then used as prompts by the Subfigure Identifier to detect subfigure bounding boxes, with the output provided as coordinates. This step results in aligned image-caption pairs that are clear and well-structured.

The extracted text is then refined by the Text Cleaner agent, which removes irrelevant information, such as book indexes and appendices, as well as any corrupted characters. Following this, the Concept Summarizer agent extracts and matches Unified Medical Language System (UMLS) concepts across multiple modalities, including text, images, tables, and diagrams. Unlike traditional text-only tools such as MedCAT~\cite{kraljevic2021multi}, the Concept Summarizer incorporates multimodal information to generate accurate and cohesive indexing concepts. By dynamically summarizing five to ten indexing concepts based on content length, this agent minimizes the risk of hallucinations that may arise from enforcing a fixed output number. The output of this stage includes clean text, image-caption pairs, tables, and diagrams, each paired with their corresponding indexing concepts to provide a unified and accessible knowledge structure.

To create a bilingual echocardiography knowledge base, the processed content from English textbooks is translated into Chinese. However, English clinical guidelines are not translated into Chinese to avoid potential conflicts arising from regional differences in guideline recommendations. The entire curation process is validated by two board-certified echocardiographers, who confirm that the source materials for the Expertise Database are authoritative and comprehensive. Additionally, they verify that the output samples of each step in the pipeline are clinically accurate and relevant, ensuring the reliability and quality of the curated knowledge.

\noindent\bn{\textbf{Integrating EKL in EchoChat's training data.} Upon constructing EKL, we integrate it into the training data by unifying knowledge across various formats using indexing concepts. Specifically, for an EKL entry that contains an image, we align its indexed concepts with those of entries featuring detailed echocardiography-related texts to retrieve enriched textual information. FAISS~\cite{douze2024faiss} is utilized during this process to identify the top-$k$ most relevant entries, which serve as background knowledge for the image. In our implementation, we retrieve the top-2 relevant entries. These retrieved data are then employed to construct various visual question answering tasks, such as image captioning, echocardiography knowledge description, and keyword summarization. These visual question answering pairs then become part of the training data. This approach fosters the model's ability to associate echocardiography images with both contextual and cardiology-specific background knowledge, thereby providing a richer understanding of echocardiography.}

\subsection{Cardiologist-Like Reasoning}

Clinically, echocardiographers follow a structured decision-making process before arriving at a well-supported and comprehensive diagnosis. This process involves observing echocardiography to identify abnormal visual clues, leveraging cardiology knowledge to propose potential disease candidates, using imaging findings and clinical knowledge to confirm or eliminate certain options, and finally drawing a conclusion. However, most current echocardiography interpretation models are trained exclusively on final diagnostic conclusions, without incorporating the reasoning process that leads to those diagnoses. This misalignment between the model's decision-making process and that of clinicians prevents echocardiographers from understanding the model's rationale or supporting evidence, leading to reduced accuracy, trust, and adoption of these models' predictions.

To bridge this gap and improve diagnostic performance, we propose the Cardiologist-Like Reasoning (CLR) framework. CLR leverages clinical reports and LLM-based cardiology knowledge to construct a comprehensive reasoning path that derives diagnostic conclusions step-by-step from echocardiographic findings. Specifically, echocardiography reports are typically composed of two key components: the findings, which document all visual observations from the echocardiography, and the summary, which contains the final conclusions about heart conditions and diseases. While these components provide critical diagnostic clues, the causal relationship between the findings and the conclusions, i.e., how specific findings lead to a particular diagnosis, is not explicitly outlined in the reports. To address this, we use a large language model (LLM) with carefully designed prompts to construct the reasoning path for CLR.

First, we utilize the echocardiography report to generate a set of positive and negative disease candidates using the LLM. Positive candidates include explicitly mentioned diagnoses and abnormalities, representing the confirmed ground-truth conditions in the report. Negative candidates are the remaining conditions in the predefined disease-label set after excluding the report-positive labels. This supervision treats clinically approved reports as a proxy for imaging findings and assumes that conditions not mentioned in the report are absent. Together, these candidates provide positive and negative targets for the structured reasoning supervision.

Building on these candidates, we prompt the LLM to associate the visual findings in the report with the disease hypotheses by applying cardiology and echocardiography diagnostic knowledge. For positive candidates, the LLM is prompted to identify related echocardiographic findings, the etiology of these findings, and the key distinguishing features of the disease. For negative candidates, the LLM provides the expected echocardiographic findings for the disease, contrasts them with the actual findings, and explains the rationale for ruling out the disease.

After reordering the logical sequence of this evidence, we construct a reasoning path that mirrors the clinician's diagnostic workflow. This path begins with observing echocardiography for visual clues, proceeds to proposing disease candidates and forming diagnostic hypotheses, incorporates echocardiographic evidence and clinical knowledge to confirm or reject these hypotheses, and concludes with the final diagnostic decision. By expressing this sequence, CLR makes the model's stated findings and diagnostic inferences available for inspection.
By capturing both affirmative and exclusionary reasoning patterns across a broad range of cases, our approach provides supervision for both supporting and exclusionary diagnostic explanations. This includes not only identifying evidence supporting a diagnosis but also systematically ruling out alternative explanations. The resulting structured records are converted into reasoning VQA pairs and trained with the corresponding images and videos. At inference, the model receives imaging inputs and a clinical question, and generates a rationale with its final answer. The recognition benefit is assessed in the ablation analysis.

\subsection{Model Design and Training}

\methodname{} is a multi-modal large language model built upon the Qwen2.5-VL-7B architecture. It consists of a 32-layer Vision Transformer (ViT) and a 7-billion-parameter decoder-only large language model. The ViT encodes the complete echocardiography study into image or video tokens, while the language tokenizer transforms echocardiography-related instructions (e.g., questions) into text tokens. These image/video tokens are then concatenated with the text tokens and fed into the large language model to predict the next token. \methodname{} generates responses in an autoregressive manner, ensuring sequential prediction.

We initialized \methodname{} with pretrained weights~\cite{bai2025qwen2} and instruction-tuned it using a large-scale multi-modal corpus. During training, the learning rate was linearly warmed up to an initial value of $1e^{-5}$ over the first 5\% steps. We utilized a batch size of 1 per device and trained the model on 32 NVIDIA H800 80 GB GPUs. The entire set of preprocessed echocardiography studies and their corresponding instructions were fed into the model for training. The model was optimized using cross-entropy loss and trained for one epoch on the curated instruction-tuning dataset to prepare it for evaluation.

For the ablation study, we randomly selected 6,000 patients from the training cohort. All other training data for these patients were held constant, while training entries from EKL or CLR were included or excluded. Performance was evaluated on the internal test set consistent with the main experiments. The aggregated results are macro-averaged across diseases.
Holm correction across 12 tests covered the three ablation configurations versus the base model across four metrics. A separate 16-test family covered component-addition comparisons across the four metrics, including the addition of EKL when CLR was already present and the addition of CLR when EKL was already present.

For evaluation, we fixed the random seed to 42. The selected, preprocessed videos and static images were fed into the model without any additional augmentation. This ensures that the evaluation process remains consistent and directly reflects the model's capabilities.
For public-benchmark LVEF evaluation, we used all 10,027 patients in EchoNet-Dynamic~\cite{ouyang2020video} as an external cohort, with the unchanged video-processing and inference protocol used for the PanEcho comparison~\cite{holste2025complete}. Each prediction take a single A4C view video as the input.
For external VQA, EchoChat was evaluated by direct inference on 622 held-out multiple-choice video questions from MIMIC-Echo-VQA~\cite{mimic-echo-vqa}, which was excluded from training, without benchmark-specific fine-tuning. Comparator results were taken from published evaluations, retaining their direct or agentic inference settings. This assessed transfer to held-out questions in a format represented during training.

Inference timing was measured on one NVIDIA H800 80~GB GPU at batch size one, excluding preprocessing. A cardiac condition-recognition query required $1.939\pm1.869$~s across 300 queries, and generation of a complete ten-section report required $7.884\pm0.959$~s across 500 queries (mean $\pm$ standard deviation). Study inputs comprised the selected 16-frame videos for the 11 main views and all static images. These timings describe model inference, separately from clinician review and report finalization.

\subsection{Cardiac disease recognition in comparison with human echocardiographers}

We evaluated the diagnostic performance of \methodname{} by comparing it against both routine and senior echocardiographers on a cohort including clinically challenging cases. A total of 100 examinations from Mainland A China, held out from Chinese report-adapter training, were selected for the study: 72 with cardiac conditions requiring careful multi-view and multi-modal echocardiographic observations and 28 without detected cardiac abnormalities. These were the same examinations used in the collaboration study. The complete raw echocardiography studies were extracted from PACS and provided to human echocardiographers for evaluation. To ensure the reliability of the ground truth conclusions, one senior echocardiographer independently reviewed and validated the correctness of the initial conclusions from clinically approved reports before the study began. This senior echocardiographer were excluded from the remainder of the study to avoid annotation bias.

We recruited two routine clinical echocardiographers and two senior echocardiographers to participate in the study.
We define routine clinical echocardiographer as the echocardiographer with less than four years of experience on performing echocardiography examination, and senior echocardiographer as the echocardiographer with more than ten years of such experience. All participating echocardiographer are qualified to issue echocardiography reports.
All routine clinical echocardiographers recruited for this reader study had similar levels of experience, with years of practice differing by less than two years, and each performed a similar number of echocardiographic examinations per year.
The anonymized echocardiography studies were randomized and provided to the echocardiographers, who were instructed to observe the studies and write echocardiography reports.

In parallel, we preprocessed the same set of echocardiography studies and used them as input to \methodname{} to generate the echocardiography report. After collecting the conclusions from all echocardiographers, we employed Qwen3.6-27B to map the ground truth, \methodname{}’s report, and the report from the echocardiographers to the same 21 clinician-defined primary diagnostic labels using the full-report extraction approach described above. Finally, we calculated and compared precision, recall and primary micro-F1 from prediction. Micro-F1 pools true positives (TP), false positives (FP) and false negatives (FN) across all patient-diagnosis pairs, so each diagnostic decision contributes equally to the score. The 21 cardiac disease label set focuses on primary diagnoses and contains 320 reference-positive patient-diagnosis pairs.

\subsection{Expert evaluation of \methodname{}-echocardiographer collaboration}

We further explored the clinical utility of \methodname{} by studying its collaboration with routine echocardiographers on the report generation task to improve the quality of echocardiography reports. The expert evaluation of \methodname{}-echocardiographer collaboration consisted of two steps: (1) \methodname{}-routine echocardiographer collaborative revision, in which routine echocardiographers refined echocardiography reports generated by \methodname{} by identifying and correcting errors, and (2) source-blinded senior echocardiographer scoring, where two senior echocardiographers independently assessed five reports per examination using six separate five-point dimensions covering clinical content and report expression.

\noindent \textbf{\methodname{}-routine echocardiographer collaborative revision.}
We used the same 100 examinations for this experiment, including 72 using the same screening criteria as in the previous section and 28 without detected cardiac abnormalities. The data exportation and ground truth screening procedures remained consistent with those described in the previous section.
Two additional routine clinical echocardiographers were recruited for the collaboration process. Using the preprocessed dataset of 100 echocardiography studies, we first generated all reports using \methodname{} with the Chinese report generation adapter trained on the other 6,757 Mainland A examinations. For each echocardiography study and the corresponding \methodname{}-generated report, they were instructed to: (1) identify errors in the AI-generated report, (2) classify the type of error and assess its clinical significance, and (3) provide corrected alternatives for the errors. After the collaborative process, the erroneous sentences in the \methodname{}-generated reports were replaced with the corrected alternatives to produce the final collaborative reports.
Editing reasons were grouped as unsupported findings, omitted findings, severity/measurement discrepancies and other reasons. Each category counted reports with at least one such edit out of 100 saved reports per collaborator, allowing overlap. Clinical-significance groups were mutually exclusive: unchanged, only edits marked non-significant, or at least one edit marked clinically significant by the revising clinician. Separately, reference-based label transitions counted omitted diagnoses added, false positives removed, correct diagnoses lost and false positives introduced, using patient-diagnosis pairs as the unit (\Cref{fig:userstudy}d-f).

\noindent \textbf{Source-blinded senior echocardiographer scoring.}
Following the collaboration process, two senior echocardiographers each with more than ten years of experience on performing echocardiography examinations, were recruited to independently score the quality of the reports. For each echocardiography study, five reports were prepared for evaluation: two collaborative reports, two reports individually written by routine echocardiographers, and one report generated solely by \methodname{}. To ensure unbiased evaluation, the sources of the reports were anonymized, and the order of the reports was randomized to eliminate potential hints during the evaluation process. Both raters assessed all five reports for each of the 100 examinations, with unified report presentation format and access to the original echocardiographic examinations.

To comprehensively assess the report's quality based on echocardiographic content faithfulness, logic coherence, and reporting language, a six-dimensional grading rubric is designed and verified by an American Society of Echocardiography (ASE) level III echocardiographer. The evaluation dimension spans imaging finding accuracy, diagnostic conclusion accuracy, measurement and severity accuracy, completeness, internal consistency, and clarity and report expression. Within each dimension, detailed assessment content and five-point scoring anchors are provided.
Scores range from 1 to 5, with higher values indicating better quality. For the first five dimensions, raters were instructed to assess clinical content without favoring narrative or structured formats. The sixth dimension, clarity and report expression, assessed language separately; its score could not offset deficiencies in clinical content.

Each senior echocardiographer independently reviewed the echocardiography studies and the anonymized reports. They were instructed to score the reports based on their quality under the six-dimensional framework and to provide brief justifications for their scoring, if necessary. After the evaluation, we averaged the two raters' scores within each examination and output for each dimension, and compared each collaboration with each standalone routine output in four examination-paired comparisons per dimension. We reported mean scores, paired differences, 95\% confidence intervals and Holm-adjusted $P$ values, retaining the six dimensions separately and treating examinations as the paired units rather than the 1,000 assessments as independent patients. This analysis provided direct insights into the impact of \methodname{}-routine echocardiographer collaboration on report quality.

\subsection{Evaluation metrics and statistics}
\label{sec:eval_metric}

For the heart condition recognition task, we evaluated the model's performance using Accuracy ($Accuracy = \frac{TP+TN}{TP+TN+FP+FN}$) and F1 score ($F1 = \frac{2 \times (Precision \times Recall)}{Precision + Recall}$, where $Precision = \frac{TP}{TP+FP}$ and $Recall = \frac{TP}{TP+FN}$).

For the reader study, primary micro-F1 pooled counts across the 100 examinations and 21 primary diagnostic labels: $F_{1,\mathrm{micro}}=2\sum TP/(2\sum TP+\sum FP+\sum FN)$. Macro-F1 is the unweighted mean of the 21 label-specific F1 scores. Foundation-model comparison averages gave equal weight to available condition-center or measurement-center entries shared by EchoChat and the comparator, excluding unsupported entries. Summaries displaying all models used their common subset. These diagnostic metrics are distinct from token-level F1, report abnormality-finding F1 and VQA factual accuracy. Generated diagnostic labels are discrete decisions and do not provide the continuous disease-specific scores required for conventional ROC analysis.

Severity recognition was evaluated at cumulative thresholds of at least mild (mild or greater versus none/normal) and at least moderate (moderate or greater versus none/mild), using precision, recall and F1. For ordinal agreement, reference and predicted grades were indexed in increasing order. Exact agreement was the proportion with identical grades; agreement within one grade was the proportion with an absolute grade difference of at most one; and the rate of errors of at least two grades was the proportion with an absolute difference of two or more. Linear-weighted Cohen's $\kappa$ used disagreement weights $w_{ab}=|a-b|/(K-1)$ for $K$ ordered grades, and was computed as $1-\sum_{a,b}w_{ab}O_{ab}/\sum_{a,b}w_{ab}E_{ab}$, where $O_{ab}$ and $E_{ab}$ were the observed and chance-expected proportions. Four-grade schemes distinguished none/normal, mild, moderate and severe; three-grade schemes combined moderate and severe. The comparison with PanEcho~\cite{holste2025complete} on aortic stenosis task in external centers combined mild and moderate to form normal, mild/moderate, and severe grades, following PanEcho's protocol. EchoChat's original four-grade outputs were retained for cumulative-threshold evaluation and detailed aortic-stenosis confusion matrices.
Baseline comparisons were made at compatible cumulative thresholds, and ordinal results were interpreted within the grade scheme specified for each condition and model. PanEcho's aortic-stenosis output merged mild and moderate disease, so its at-least-moderate endpoint was unsupported.

For the heart measurement estimation task, we used Mean Absolute Error (MAE, $MAE = \frac{1}{n} \sum_{i=1}^n \left| y_i - \hat{y}_i \right|$) and $R^2$ score ($R^2 = 1 - \frac{\sum_{i=1}^n \left( y_i - \hat{y}_i \right)^2}{\sum_{i=1}^n \left( y_i - \bar{y} \right)^2}$) to assess the model's performance. LVEF errors were expressed in \%. Scatter plots compared predictions with reference LVEF using the identity line. Bland-Altman plots displayed prediction-minus-reference differences against their paired means, with mean bias and 95\% limits of agreement defined as mean bias $\pm1.96$ times the standard deviation of the differences.

For echocardiography report generation, we employed Natural Language Generation (NLG) metrics, including the Bilingual Evaluation Understudy (BLEU) score~\cite{papineni2002bleu}, Recall-Oriented Understudy for Gisting Evaluation (ROUGE) score~\cite{lin2004rouge}, and Metric for Evaluation of Translation with Explicit Ordering (METEOR) score~\cite{banerjee2005meteor}.
For the echocardiography visual question answering task, we also used NLG metrics, including BLEU score~\cite{papineni2002bleu}, ROUGE score~\cite{lin2004rouge}, sentence recall, and F1 score. Sentence recall and F1 were computed at the token level by comparing predicted sentence tokens with ground truth tokens.

To ensure robustness, we further calculated the 95\% confidence interval (95\% CI) for recognition, measurement and language-evaluation metrics using bootstrapping with 1000 iterations.
For the reader study, patient-bootstrap confidence intervals with 5000 iterations retained all outputs from each examination together. Quality-score comparisons first averaged the two raters within each examination and output, separately by dimension. The four collaboration-versus-routine comparisons were C1/C2 versus Routine A/B; the diagnostic gain plot also included comparisons with EchoChat alone (\Cref{fig:userstudy}b). Significance was reported using Holm-adjusted $P$ values. The separate human-LLM VQA agreement analysis used 20,000 patient-cluster bootstrap replicates.

The EKL/CLR ablation compared four configurations: the base model without either component, EKL alone, CLR alone and both components. Accuracy, precision, recall and F1 were compared using two-sided paired patient-level permutation tests. Holm correction across 12 tests covered the three augmented configurations versus the base model across four metrics. A separate 16-test family covered component-addition comparisons across the four metrics, including the addition of EKL when CLR was already present and the addition of CLR when EKL was already present.

\noindent \textbf{Abnormality-finding precision, recall, and F1 metric for report generation quality evaluation.}
To assess abnormality-concept agreement in generated echocardiography reports while reducing the influence of reporting style, we compute abnormality-finding precision, recall, and F1 using a report-processing pipeline similar to that of NeuroVFM~\cite{kondepudi2026health}. Precision measures the proportion of generated findings supported by the reference report, penalizing unsupported findings. Recall measures the proportion of reference findings captured in the generated report, reflecting the completeness of the generated content. F1 combines precision and recall as their harmonic mean.
These metrics assess abnormality-concept agreement and coverage; severity and measurement accuracy are evaluated separately.

For each reference and generated report, we first use the Qwen3.6-27B large language model (LLM) to extract echocardiographic findings through a three-step prompting procedure. This procedure excludes statements describing normal findings, recommendations, comparisons, dates, and clinical history, while retaining trace and mild abnormalities. We then use the same LLM to match the extracted findings between the generated and reference reports, accounting for synonyms and semantically equivalent expressions. Independent findings are atomized and clinically equivalent concepts deduplicated, with a qualitative finding and its supporting measurement retained as one concept. Matching requires compatible abnormality identity and anatomy, including chamber, valve, leaflet or wall segment and disease, procedure or device state. Severity differences do not prevent a match when the underlying abnormality and anatomy agree; differences that change finding presence or clinical identity do. Each generated concept matches at most one reference concept, while multiple generated concepts may match the same reference concept.

For report pair $i$, let $\mathrm{G}_i$ and $\mathrm{P}_i$ denote the sets of reference and generated findings, respectively. We define the matching function $m_i:\mathrm{P}_i \rightarrow \mathrm{G}_i \cup\{\varnothing\}$, where $m_i(p)=\varnothing$ indicates that generated finding $p$ is unsupported by the reference report. The number of supported generated findings, $S_i$, and the number of distinct reference findings covered by the generated report, $C_i$, are defined as
\begin{equation}
S_i = \sum_{p \in \mathrm{P}_i} \mathbf{1}\!\left[m_i(p)\neq\varnothing\right],
\qquad
C_i = \left|\left\{m_i(p):p\in\mathrm{P}_i,\ m_i(p)\neq\varnothing \right\}\right|.
\end{equation}
For nonempty reference and generated finding sets, abnormality-finding precision and recall are
\begin{equation}
P_i=\frac{S_i}{|\mathrm{P}_i|},
\qquad
R_i=\frac{C_i}{|\mathrm{G}_i|}.
\end{equation}
Abnormality-finding F1 is their harmonic mean:
\begin{equation}
F_{1,i}=
\begin{cases}
\dfrac{2P_iR_i}{P_i+R_i}, & P_i+R_i>0, \\
0, & P_i+R_i=0.
\end{cases}
\end{equation}
When both concept sets are empty, precision, recall, and F1 are all assigned a value of $1$. When exactly one finding set is empty, all three metrics are assigned a value of $0$. Because matching is many-to-one, multiple generated findings may match the same reference finding; consequently, $S_i$ and $C_i$ can differ, and precision and recall use distinct numerators. We report aggregate performance by averaging each metric across individual report pairs.

\noindent \textbf{Factual accuracy metric for visual question answering faithfulness evaluation.}
We compute factual accuracy to quantify how faithfully EchoChat responses preserve the findings in the corresponding echocardiographic contents. Specifically, we used Qwen3.6-27B as a large language model (LLM) judge to assess agreement between each model response and a reference answer derived from the echocardiography report. The judge receives the question, reference answer, and generated response, and returns a binary consistency decision with a brief justification. The question and reference answer serve as the sole basis for evaluation, without introducing external medical knowledge or unstated clinical thresholds. The evaluation focuses on whether the response conveys the core factual information requested by the question. Factual accuracy is calculated separately for each question category as
\begin{equation}
\text{Factual Accuracy}
=
\frac{1}{N}\sum_{i=1}^{N} c_i,
\end{equation}
where $N$ is the number of evaluated responses in that category and $c_i=1$ if the judge determines that response $i$ is consistent with its reference under the specified rubric, and $c_i=0$ otherwise.

To evaluate agreement between the LLM judge and human raters, two American Society of Echocardiography (ASE) level II echocardiographers assessed the concordance between EchoChat responses and reference answers for 240 visual question answering items, equally sampled across eight question topics, using the same evaluation instructions as the LLM judge. We quantified pairwise agreement using percentage agreement and Cohen's $\kappa$, with 95\% percentile confidence intervals estimated using 20,000 patient-cluster bootstrap replicates. The LLM judge agreed with rater~1 on 226/240 items (94.2\%; $\kappa=0.786$, 95\% CI, 0.671-0.883) and with rater~2 on 206/240 items (85.8\%; $\kappa=0.536$, 95\% CI, 0.394-0.662). Both LLM-human comparisons showed numerically higher percentage agreement and $\kappa$ than the human-human comparison (202/240 items, 84.2\%; $\kappa=0.464$), providing a human agreement benchmark for interpreting the LLM's performance. Among the 202 items on which both echocardiographers agreed, the LLM matched their shared judgment on 197 (97.5\%). Together, these findings demonstrate agreement beyond chance with both echocardiographers and close alignment with their shared judgments, supporting the use of the LLM judge to evaluate response-reference concordance in this study.

\section{Data availability}
The dataset of echocardiography studies and reports used to train and validate \methodname{} are the proprietary data of the participating hospitals, hence will not be publicly available.

\section{Code availability}
The code and model developed in this study will be publicly released upon publication of the manuscript. Details of the implementation, including the network architecture and training strategies, are provided in the Methods section.

\section{Acknowledgments}

The work described in this paper was supported by grants from the Research Grants Council (RGC) of the Hong Kong Special Administrative Region, China (Project Nos. R6005-24, N\_HKUST654/24, and C6088-25Y). It was also partially supported by RGC research grants under Project Nos. AoE/E-601/24-N and T45-401/22N, a research grant from the Innovation and Technology Commission (ITC) (Project No. GHP/124/22), a research grant from the National Natural Science Foundation of China (Grant No. 62622217, 82371963, and 82672514), and a research grant from Guangdong Basic and Applied Basic Research Foundation (Grant No. 2025A1515220120).
The EU validation is part of the DVPS project (Diversibus Viis Plurima Solvo) supported by the European Union's Horizon Europe research and innovation programme under grant agreement No 101213369. The Spanish validation was funded by the Spanish Ministry of Science, Innovation and Universities (PID2021-128367OA-I00 and FORT23/00034) and the Instituto de Salud Carlos III (CP24/00121).
The authors thank Qiu Xie, Bo Kong, Lei Huang, Yu Wang, Jiexuan Zheng, Jinxin Xu, Xuan Sha, and Yuan Zhu from Guangdong Provincial People’s Hospital for participating in the echocardiographer reader study.
The authors also thank the Hong Kong Generative AI Research and Development Center (HKGAI) for providing partial support for the computing resources used in this study.

\section{Author contributions statement}

X.M.-L. designed the study.
Y.Q., M.E., and S.S. developed and implemented the proposed method, benchmarked different AI models, and analyzed the results. Specifically, Y.Q. and S.S. focused on developing the new methods, while M.E. was responsible for the training pipeline and for benchmarking the models across external testing hospitals.
T.H. provided clinical input on the study design, and helped design and organize the user study.
H.F. provided the Chinese external testing data, provided clinical input on the study design, and helped design the user study.
J.G. helped raw data preprocessing and model deployment for testing.
L.W. contributed to raw data preparation and experimental design for visual question answering tasks.
N.L. contributed to raw data preparation.
C.Z. and A.PW.L. provided training data and provided clinical opinion in study design.
A.M., M-C.V., F.A., and K.L. provided European external testing data.
E.A-M. and S-J.R. helped processing the Austrian data and inferring the models in Austria.
M-J.B. and S.Z. helped organizing and processing the Netherlands data and inferring the models in Netherlands.
A.G., L.D.-S., P.L.-G. helped organizing and processing the Spainish data and inferring the models in Spain.
Y.G. provided partial computational resource support for this study.
Y.Q. and S.S. drafted the manuscript, with contributions from M.E. and revisions by A.G., L.D.-S., P.L.-G., M.C.-V., F.A., N.J.S., H.Z, and X.M.-L.
A.P-C., Y.Z., D.T., R.D., N.J.S., and H.Z. analyzed the results, and improved the revised manuscripts. All authors discussed the results and contributed to the final manuscript. H.W.-F., A. PW-L., X.M.-L. are the corresponding authors.

\bibliographystyle{splncs04}
\bibliography{bib}
\end{document}